\documentclass[10pt,journal,compsoc]{IEEEtran}
\usepackage{cite}
\usepackage{amsmath,amssymb,amsfonts}
\usepackage{algorithm}
\usepackage{algorithmicx}
\usepackage{algpseudocode}
\usepackage{graphicx}
\usepackage{textcomp}
\usepackage{epsfig}
\usepackage{graphics}
\usepackage{color}
\usepackage{subfigure}
\usepackage{multirow}
\usepackage{threeparttable}
\usepackage{booktabs}
\usepackage{multirow}
\usepackage{soul}
\usepackage{epstopdf}
\usepackage{psfrag}
\usepackage{setspace}
\usepackage{subfigure}
\usepackage{tabularx}
\usepackage{makecell}
\usepackage{arydshln} 
\usepackage{booktabs}
\usepackage{multirow}
\usepackage[hyphens]{url}
\usepackage{hyperref}
\usepackage{array}
\usepackage{adjustbox}

\newcommand{\REO}[1]{\textcolor{black}{#1}}
\newcommand\etal{\emph{et~al.}}
\hypersetup{
 linkcolor=green,
 anchorcolor=green,
 citecolor=black,
}

\begin{document}

\title{Exploring Cognitive and Aesthetic Causality \\for Multimodal Aspect-Based Sentiment Analysis}

\author{Luwei Xiao,~\IEEEmembership{Student Member,~IEEE},
    Rui Mao$^*$,~\IEEEmembership{Member,~IEEE},
    Shuai Zhao,\\
    Qika Lin,
    Yanhao Jia,
    Liang He,
    and Erik Cambria,~\IEEEmembership{Fellow,~IEEE}
\IEEEcompsocitemizethanks{\IEEEcompsocthanksitem Luwei Xiao, and Liang He are with the School of Computer Science and Technology, East China Normal University, Shanghai 200062, China. E-mail: louisshaw@stu.ecnu.edu.cn, lhe@cs.ecnu.edu.cn

\IEEEcompsocthanksitem Rui Mao, Shuai Zhao, Yanhao Jia and Erik Cambria are with the College of Computing and Data Science, Nanyang Technological University, Singapore 639798. E-mail:\{rui.mao, shuai.zhao, cambria\}@ntu.edu.sg, yanhao002@e.ntu.edu.sg

\IEEEcompsocthanksitem Qika Lin is with the Saw Swee Hock School of Public Health, National University of
Singapore 119077. E-mail: linqika@nus.edu.sg

}
\thanks{* Corresponding author: Rui Mao}}

\markboth{IEEE TRANSACTIONS on Affective Computing,~Vol.~XX, No.~XX, XXXX~2025}%
{Shell \MakeLowercase{\textit{et al.}}: Bodily Expressed Emotion Recognition Framework Based on Spatio-Temporal Features}

\IEEEtitleabstractindextext{%
\begin{abstract}

Multimodal aspect-based sentiment classification (MASC) is an emerging task due to an increase in user-generated multimodal content on social platforms, aimed at predicting sentiment polarity toward specific aspect targets (i.e., entities or attributes explicitly mentioned in text-image pairs). Despite extensive efforts and significant achievements in existing MASC, substantial gaps remain in understanding fine-grained visual content and the cognitive rationales derived from semantic content and impressions (cognitive interpretations of emotions evoked by image content). In this study, we present Chimera: a \underline{\textbf{c}}ognitive and aest\underline{\textbf{h}}et\underline{\textbf{i}}c senti\underline{\textbf{me}}nt causality unde\underline{\textbf{r}}st\underline{\textbf{a}}nding framework to derive fine-grained holistic features of aspects and infer the fundamental drivers of sentiment expression from both semantic perspectives and affective-cognitive resonance (the synergistic effect between emotional responses and cognitive interpretations). Specifically, this framework first incorporates visual patch features for patch-word alignment. Meanwhile, it extracts coarse-grained visual features (e.g., overall image representation) and fine-grained visual regions (e.g., aspect-related regions) and translates them into corresponding textual descriptions (e.g., facial, aesthetic). Finally, we leverage the sentimental causes and impressions generated by a large language model (LLM) to enhance the model's awareness of sentimental cues evoked by semantic content and affective-cognitive resonance. Experimental results on standard MASC datasets demonstrate the effectiveness of the proposed model, which also exhibits greater flexibility to MASC compared to LLMs such as GPT-4o. We have publicly released the complete implementation and dataset at \url{https://github.com/Xillv/Chimera}

\end{abstract}

}

\maketitle

\IEEEdisplaynontitleabstractindextext

\IEEEpeerreviewmaketitle

\IEEEraisesectionheading{\section{Introduction}\label{sec:introduction}}

\IEEEPARstart{M}{ultimodal} aspect-based sentiment classification (MASC) is a valuable task for analyzing user-generated multimodal content on social platforms, aiming to predict the sentiment polarity of a specific target/aspect term within a sentence, based on an image-text pair. In an era marked by growing global interconnectedness, social platforms have become essential channels for individuals to express opinions and share experiences~\cite{mao2023biases,du2024financial,mao2025survey}. These platforms support multimodal content, blending text and visual media, which better reflects how sentiment is conveyed~\cite{xiao2022exploring}. Consequently, analyzing fine-grained sentiment expression in multimodal scenarios not only improves the depth of sentiment classification but also aligns with the natural manner in which users express opinions and emotions, ultimately supporting more accurate sentiment analysis for applications in finance~\cite{ma2024quantitative, du2023finsenticnet}, social research~\cite{zhang2024senticvec,zhao2024universal}, and human-computer interaction~\cite{zhu2024neurosymbolic,zhao2023prompt}.
Current methodologies for MASC can be broadly divided into two principal categories: visual-text fusion-based approaches and translation-based approaches. Visual-text fusion-based methods address MASC by directly integrating visual content with textual features through various attention-based mechanisms~\cite{yuadapting, yu2019entity, yu2022targeted, ling2022vision, yang2022cross, zhou2023aom}. 

Yu~\etal~\cite{yuadapting} were the first to propose the utilization of ResNet for image feature extraction in conjunction with BERT for language sequence modeling, subsequently feeding these components into a BERT encoder to facilitate the interactive modeling of cross-modal representations. Ling~\etal~\cite{ling2022vision} introduced a vision-language pre-training framework that leverages Faster R-CNN for extracting object-level visual features and BART for generating textual features, with the model pre-trained using three task-specific strategies targeting the language, vision, respectively. Yu~\etal~\cite{yu2022targeted} presented a novel multi-task learning framework Image-Target Matching Network (ITM), which concurrently performs coarse-to-fine-grained visual-textual relevance detection and visual object-target alignment through cross-modal Transformers. Translation-based approaches focus on mapping visual content into the language space as auxiliary textual representations, leveraging this supplementary information, or integrating it with visual features to enhance MASC~\cite{khan2021exploiting, xiao2022adaptive, xiao2023cross, huang2023target, wang2023image, xiao2024atlantis}. Khan~\etal~\cite{khan2021exploiting} translated the image into a corresponding caption, which is then jointly input with the sentence into BERT to predict the sentiment polarity associated with specific targets. Yang~\etal~\cite{yang2022face} exploit a face-sensitive, translation-based approach that translates facial expressions in images into textual sentiment cues, which are then selectively aligned and fused with the targets for enhanced sentiment analysis. Xiao~\etal~\cite{xiao2023cross} proposed the CoolNet framework, which generates visual captions for images and extracts syntactic and semantic features from the textual modality, subsequently fusing these with visual features through a cross-modal Transformer.\\

Despite substantial efforts and promising advancements, current solutions continue to encounter the following challenges. First, excessive duplicative visual patches can overshadow critical visual clues relevant to the specific target, leading to considerable misalignment during patch-token interactions. These small visual patches often lack semantic coherence compared to complete visual regions, particularly when aligning targets with their corresponding objects in an image, potentially leading to ambiguous semantic representations. Second, limited studies have focused on the underlying rationale behind sentiment cues, particularly from the perspectives of semantic content and affective-cognitive resonance. Owing to the multimodal nature of Twitter content, which spans diverse facets of daily life, inferring the sentiment associated with specific targets necessitates not only an understanding of the surface-level information in text and images (e.g., facial expressions) but also an in-depth comprehension of the contextual background of particular events and the impressions evoked by the image's content and aesthetic attributes. 

To address the aforementioned challenges, this paper proposes Chimera: a \underline{\textbf{c}}ognitive and aest\underline{\textbf{h}}et\underline{\textbf{i}}c senti\underline{\textbf{me}}nt causality unde\underline{\textbf{r}}st\underline{\textbf{a}}nding framework. This framework aims to incorporate and align fine-grained features of specific targets and reasons about semantic and impression rationales. However, two critical issues must be resolved to achieve these objectives: 1) How can specific targets in a sentence be aligned with their corresponding object-level fine-grained features in an image? 2) How can the model be enabled to reason about the emotional causal reasons within the semantic content of image-text pairs and the affective resonance evoked by image aesthetic attributes? For the first question, we propose to make the cross-modal alignment of the target via the visual patch-level by linguistic-aware patch-token alignment and object-level by accurately translating the object feature into language space. Regarding the second issue, while a recent study~\cite{fan2024dual} developed a reasoning dataset for MASC, this dataset primarily explains the emotional causes within textual content and lacks reasoning capabilities for visual content and the affective resonance evoked by images, limiting its suitability for the multimodal nature of this task. Consequently, we employ a large language model (LLM), GPT-4o, to generate the semantic rationale and impression rationale to understand the causal foundations of emotions.

Specifically, our proposed framework first extracts visual patch-level and textual features, feeding them into a tailored linguistic-aware patch-token alignment (LPA) module to achieve patch-token alignment. Concurrently, a translation module (TM) translates the holistic image or object-level content into aesthetic captions or facial descriptions, leveraging multimodal named entity annotations from the work of Wang~\etal~\cite{wang2023fine}. The TM-generated text, along with the sentence and aspect, is then input into a generative module for multi-task learning to produce sentiment polarity, semantic rationale (SR), and impression rationale (IR). By bootstrapping the model's perception of underlying rationale through an in-depth understanding of textual and visual content as well as the affective resonance evoked by images, it enhances the performance of sentiment classification. 

In a nutshell, the primary contributions are as follows:
\begin{itemize}
	\item We propose a novel framework for MASC that aligns specific targets with their corresponding visual objects at the patch-token and object levels while equipping the model with causal rationale reasoning ability for semantic rationale (SR), and impression rationale (IR).
	\item We approach this task by enabling the model to grasp the semantic content of image-text pairs and the affective resonance evoked by images. To our knowledge, we are the first to collect semantic and impression rationale data for the MASC task, based on existing MASC datasets, extending its content to incorporate semantic and impression rationale, offering a valuable resource for advancing MASC research.
	\item Experiments on three widely-used Twitter benchmarks demonstrate that our proposed method outperforms previous approaches, achieving state-of-the-art performance. Further evaluations validate the effectiveness of our approach for MASC tasks.
\end{itemize}

\begin{figure*}[t!]
	\centering
	\centerline{\includegraphics[width=0.98\textwidth]{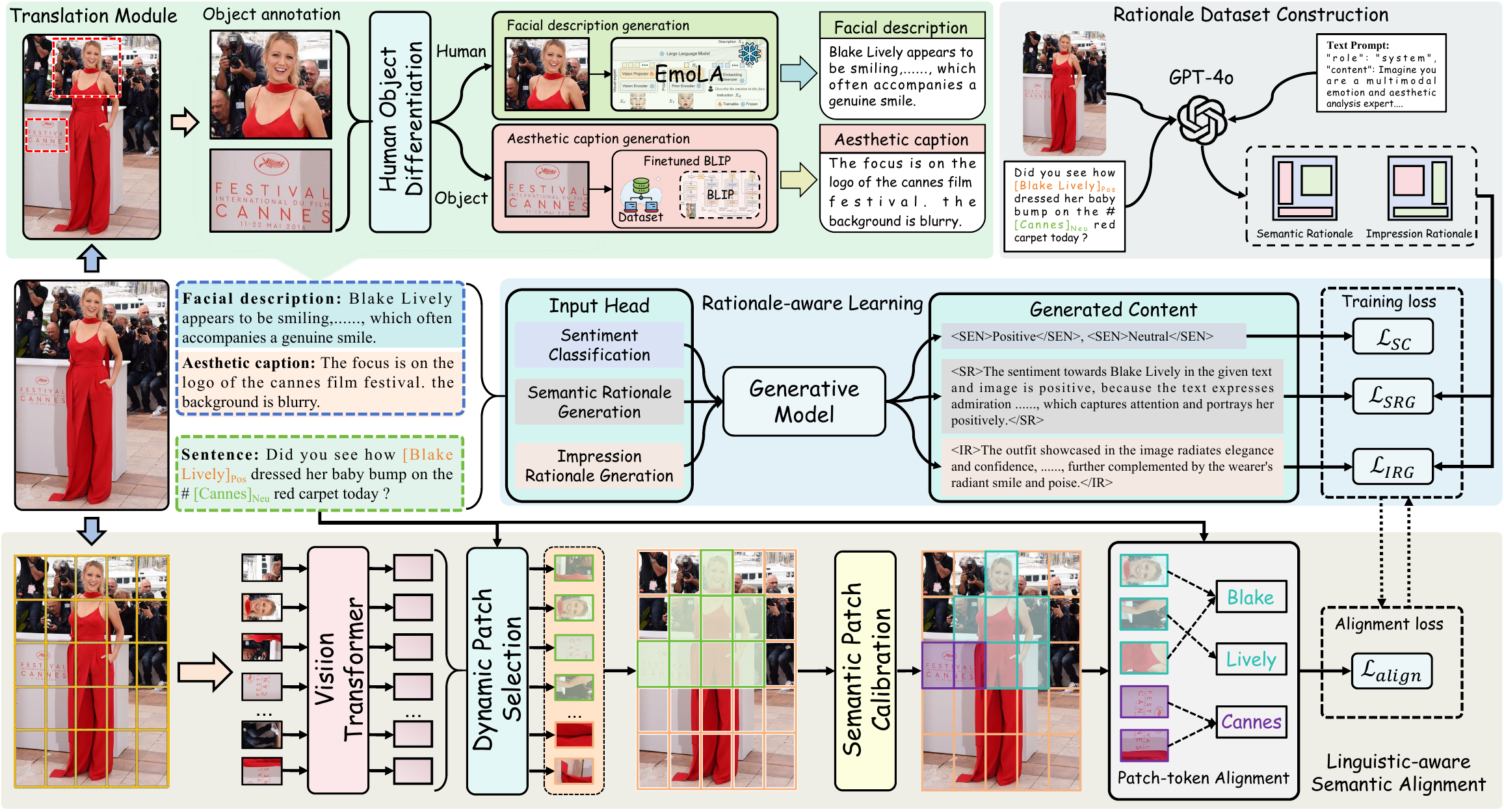}}
	\caption{The overall framework of the proposed Chimera. Chimera consists of four parts: Translation Module, Rationale Dataset Construction, Linguistic-aware Semantic Alignment, and Rationale-Aware Learning.}
	\label{fig1}
\end{figure*}

The remainder of this paper is organized as follows: Section \ref{sec:relatedwork} provides an overview of related research on multimodal aspect-based sentiment classification, image aesthetic assessment, and multimodal learning. Section \ref{sec:method} details the proposed framework, including linguistics-aware patch-token alignment, the translation-based module, causal rationale dataset construction, and LLM-based annotation generation. Main experimental results are presented in Section \ref{sec:result}, and the in-depth analysis is shown in~\ref{in:analysis}, followed by conclusions in Section \ref{sec:conclusion}.

\section{Related Work}\label{sec:relatedwork}
This section reviews key methods in multimodal aspect-based sentiment analysis and image aesthetic assessment. Additionally, as our novel rationale dataset is constructed using an LLM, we introduce LLMs for data annotation.

\subsection{Multimodal Aspect-based Sentiment Analysis}
Sentiment analysis is a well-established research area focused on understanding and identifying human emotions and opinions across various contexts~\cite{zhang2023neuro, lu2023sentiment, liu2023interpretable, mao2023discovering,cambria2024senticnet,du2024explainable}. 
With the exponential growth of user-generated multimodal content (e.g., image-text pairs, video clips) on social platforms~\cite{zhang2024privfr, yang2024model, xiao2024vanessa, kruk2019integrating} has drawn substantial attention to Multimodal Aspect-based Sentiment Analysis (MABSA)~\cite{liu2022towards, mao2021bridging,yue2023knowlenet,fan2024fusing, yang2024empirical}. 
The MABSA task consists of two sub-tasks: Multimodal Aspect Term Extraction (MATE) and our focused MASC task. MATE~\cite{yang2023generating} is essentially a named entity recognition task aimed at identifying all relevant specific targets within the textual content of an image-text pair. MASC~\cite{zhou2021masad, zhang2022survey} is a text classification task in which specific targets are provided, requiring the identification of their sentiment polarity (positive, neutral, or negative) based on the given image-text pair. A series of recent studies have successfully unified these two subtasks into a single framework, effectively streamlining the MABSA process~\cite{ju2021joint,yang2022cross,ling2022vision,mu2023mocolnet,zhao2023m2df,xiao2024atlantis,7pillars}.
Among these studies, Yu~\etal~\cite{yu2019entity} proposed the Entity-Sensitive Attention and Fusion Network (ESAFN), which employs entity-oriented attention combined with a visual gate mechanism to model entity-sensitive inter-dynamics for MASC.
Ju~\etal~\cite{ju2021joint} were the first to integrate MATE and MASC into a end-to-end task, developing a joint learning framework with cross-modal relation detection. 
\REO{Kruk~\etal~\cite{kruk2019integrating} proposed a multimodal framework for Instagram intent detection, integrating three taxonomies and the MDID dataset. It demonstrates that text-image fusion enhances accuracy by 9.6\% under semiotic divergence, emphasizing the necessity of multimodal models for capturing the non-intersective "meaning multiplication" inherent in social media.}
Yang~\etal~\cite{yang2022cross} improved cross-modal alignment modeling through a Transformer-based multi-task learning framework, incorporating text-guided cross-modal interactions and using adjective–noun pairs as supervision for a visual auxiliary task. 

Zhou~\etal~\cite{zhou2023aom} developed an aspect-oriented multimodal fusion approach that constructs an informative dependency graph to minimize additional visual and textual noise in cross-modal interactions by selectively processing aspect-relevant textual and image features. Huang~\etal~\cite{huang2023target} put forward to mapping images into scene graphs, using triplet semantic relationships among entities along with image captions to construct a relatedness matrix for achieving cross-modal alignment in MASC. More recently, Xiao~\etal~\cite{xiao2024atlantis} introduced the Atlantis, a trident-shaped architecture that incorporates aesthetic attributes to enhance the emotional resonance of visual content. Fan~\etal~\cite{fan2024dual} devised a Flan-T5-based multi-task learning architecture to enhance the model's reasoning capabilities for inferring underlying and direct causes of sentiment expressions. Additionally, they constructed a practical causal dataset for MASC.

Our proposed method aims to achieve cross-modal alignment at the patch and object levels while equipping the model with reasoning capabilities to discern the semantic and impression rationale underlying sentiment expressions.

\subsection{Image Aesthetic Assessment}

Image aesthetics play a fundamental role in shaping viewers' emotional responses and overall aesthetic experience through complex psychological and cognitive processes~\cite{arnheim1954art}. Image aesthetics pertain to the subjective evaluation and appreciation of its beauty~\cite{ramachandran1999science}. Image Aesthetic Assessment seeks to systematically appraise this aesthetic quality by analyzing the visual appeal of images~\cite{zeng2019unified}. Empirical psychological research corroborates that images can trigger a wide range of emotions, which are influenced by their aesthetic attributes and semantic content~\cite{cupchik1992emerging}. Previous research concentrated on aesthetic image captioning and analysis through the aggregation of commentary on aesthetic attributes~\cite{jin2019aesthetic}. 
These studies address the concepts of style, layout, and aesthetics from the viewpoints of beauty and visual attractiveness. Recent scholarly efforts have focused on encouraging vision-language models to generate visual connotations and captions related to various aesthetic attributes (e.g., color, harmony, lighting, composition)~\cite{ke2023vila}. More recently, Kruk~\etal~\cite{kruk2023impressions} introduced a connotation-rich dataset, Impressions, designed to explore the emotions, thoughts, and beliefs that images evoke, along with the aesthetic elements that elicit these responses. The introduction of this dataset marks a significant advance in the study of how visual stimuli can influence complex perceptual and emotional outcomes. 

In this study, we utilize aesthetic attributes to capture sentiment cues within visual content at both object and holistic levels. Inspired by Impressions~\cite{kruk2023impressions}, we further prompt the LLM to generate impression rationales for MASC, enabling analysis of the underlying affective resonance evoked by images.

\subsection{LLMs-Based Rationale Generation}
Recently, LLMs have achieved significant success across various downstream tasks~\cite{anil2023palm,mao2024gpteval,zhao2024exploring,zhao2025clean}. LLMs such as GPT-4o~\cite{achiam2023gpt}, Gemini~\cite{team2023gemini}, and LLaMA-2~\cite{touvron2023llama} hold significant potential to usher data annotation into a new era, functioning not merely as auxiliary tools but as vital enhancers of its effectiveness and quality~\cite{liu2023robust,mao2024metapro}. LLMs can automatically annotate samples, ensure consistency across large data volumes, and adapt to specific domains via fine-tuning, thereby establishing a new standard in deep learning~\cite{tan2024large,mao2025comparative,jia2025uni}. The rationale represents the detailed cognitive process an individual typically follows when solving a problem, providing useful supplementary information for the final answer~\cite{wei2022chain}. Early studies~\cite{cobbe2021training} typically relied on human experts to annotate rationale in datasets, significantly limiting availability and scalability. A bunch of diverse methodologies have been developed to produce high-quality and fine-grained rationale. Wang~\etal~\cite{wangpinto} proposed to elucidate each choice in a sample by generating choice-specific rationales via LLMs.
Wang~\etal~\cite{wang2023scott} enhanced the credibility of generated rationales by incorporating gold-standard answers and using contrastive decoding algorithms.
Liu~\etal~\cite{liu2023logicot} laid much emphasis on curating high-quality prompts to obtain fine-grained rationales from GPT-4o and build a logical chain-of-thought instruction-tuning dataset. 
More recently, Kang~\etal~\cite{kang2024knowledge} developed a sophisticated neural reranking mechanism to dynamically retrieve highly relevant supplementary documents for generating high-quality rationales in knowledge-intensive reasoning tasks.

In this paper, we build upon the work of Wang~\etal~\cite{wang2023scott} by fully utilizing the dataset's gold-standard annotations to generate semantic and impression rationales through meticulously designed prompts. This approach ensures high-quality rationale generation while avoiding additional costs from trial-and-error OpenAI API usage fees.

\section{Methodology}\label{sec:method}
This section presents our proposed framework for MASC, beginning with the task formalization, followed by the rationale dataset construction process, and concluding with the proposed method, comprising linguistic-aware semantic alignment, a translation module, rationale dataset construction and a rationale-aware learning framework.

\subsection{Task Definition}
\label{sec:task-def}
Given a multimodal dataset $M$, each sample $X_i$ consists of an image $V_i$ paired with a sentence $S_i$ containing one or more specific targets $T_i$. The goal of MASC is to predict the sentiment polarity $ Y_i \in \{\text{Positive, Negative, Neutral}\} $ for a specific target $T_i$. Moreover, our framework infers both semantic rationale $SR_i$ and impression rationale $IR_i$, explaining the sentiment prediction $Y_i$ for a specific target $T_i$, based on multimodal semantic meaning and the affective resonance evoked by the image. In this study, the model outputs $SR_i, IR_i, Y_i$ for an input sample \(X_i = (S_i, V_i, T_i)\), where \(SR_i\) and \(IR_i\) offer supplementary sentimental cues for sentiment prediction $T_i$.

\subsection{Method Overview}
\label{sec:model-view}
As shown in Figure~\ref{fig1}, our proposed framework comprises four technical components, namely a Translation Module, Rationale Dataset Construction, Linguistic-aware Semantic Alignment, and Rationale-Aware Learning. The Translation Module converts visual content, both holistic and object-level, into language captions. For entire images, it generates emotion-laden aesthetic captions using our fine-tuned BLIP. For object-level content, it maps visuals to facial descriptions or aesthetic captions with rich emotional cues via EmoLA or our fine-tuned BLIP. The construction of the rationale dataset involves generating semantic and impression rationales. We curate prompts tailored to each rationale category and input them, along with the samples, into GPT-4o to collect the desired rationales. The Linguistic-aware Semantic Alignment module segments the input image into patches, dynamically selects and refines relevant visual patches, and achieves patch-token alignment guided by linguistic features from the input sentence. Lastly, we propose a Rationale-Aware Learning framework built up on a generative model that simultaneously learns sentiment classification, semantic rationale generation, and impression rationale generation from diverse textual inputs, such as sentences, aesthetic captions, and facial descriptions.

\subsection{Translation Module}
\REO{
This module translates visual content into overall aesthetic captions, object-level facial descriptions, or object-level aesthetic captions in textual form, embedding rich sentimental cues to facilitate object-level sentiment alignment. Specifically, we leverage object annotations from the Fine-Grained Multimodal Named Entity Recognition (MNER) task~\cite{wang2023fine}, which annotates specific targets in the sentence and their corresponding objects in the image. The MNER dataset is derived from the same Twitter dataset as the MASC datasets, incorporating the original image-text pairs from MASC. We further pre-process the MNER dataset and transfer its object annotations to the MASC dataset. To generate aesthetic captions rich in sentimental cues, we fine-tune a BLIP model using the recent aesthetic-specific dataset, \textit{Impression}~\cite{kruk2023impressions}. For facial description, we deploy the LLM-based EmoLA~\cite{li2025facial} to interpret fine-grained human mental states from images. 
}

\REO{
To tackle the challenge of potential one-to-many annotation scenarios, wherein multiple visual objects correspond to a specific target in the sentence, we calculate the similarity between the entire image and all object annotations, retaining only the object with the highest similarity score for each specific target. Subsequently, we generate various textual auxiliary sentences, based on object annotations. Firstly, in cases where the object corresponding to a specific target is absent from the image, a fine-tuned BLIP model is applied to generate an overall aesthetic caption $A^c=\left(a^c_1, a^c_2, \ldots, a^c_{N_c}\right)$ for the entire image:
\begin{equation}
A^c=BLIP_{\textit{fine}}(V),
\end{equation}
where $BLIP_{\textit{fine}}(\cdot)$ is the fine-tuned BLIP over \textit{Impression} dataset. If the object corresponding to a specific target is present in the image, we develop a Human-Object Differentiation (HOD) module based on the Sample and Computation Redistribution for Efficient Face Detection (SCRFD)~\cite{guosample} framework. 
This module determines the presence of a face within the annotated object-level visual content and assigns a facial binary label:
\begin{equation}
Y^{o_j}_i=HOD(V^{o_j}_i),
\end{equation}
where $Y^{o_j}_i \in [1,0]$ indicates whether the object-level visual content contains a face (0 for no face, 1 for face detected), and $V^{o_j}_i$ denotes the $j$-th object-level visual content in the $i$-th image. Subsequently, we generate facial descriptions or aesthetic captions for object-level visual content based on the facial binary label:
\begin{equation}
A^o=
\begin{cases}
EmoLA(V^{o_j}_i), & \text{if $Y^{o_j}_i=1$ },\\
BLIP_{\textit{fine}}(V^{o_j}_i), & \text{otherwise},
\end{cases}
\end{equation}
where $A^o=\left(a^o_1, a^o_2, \ldots, a^o_{N_o}\right)$ is the generated auxiliary sentence (facial description or aesthetic caption) for the object-level visual content. 
}
\subsection{Rationale Dataset Construction}
The current MASC benchmark includes only specific target (aspect) labels within the image-text pair sentences and their corresponding sentiment polarities. Recently, Fan~\etal~\cite{fan2024dual} introduced a dataset for MASC with cause analysis, focusing exclusively on textual semantics rather than integrating both visual and textual cues.
Moreover, they overlook the affective resonance evoked by image aesthetic attributes, eliminating a crucial layer of emotional cues and resulting in an incomplete sentiment representation. This omission hinders the holistic integration of textual and visual modalities, leading to suboptimal sentiment modeling. Therefore, we employ GPT-4o to generate semantic and impression rationales, with the detailed generation process outlined in Algorithm~\ref{algorithm：dataset}.

\begin{algorithm}[!ht]
 \caption{Rationale Dataset Construction}
 \label{algorithm：dataset}
 \begin{algorithmic}[1]
  \Require
   All samples ($V$, $S$, $T$, $Y$) in MASC dataset $M$
  \Ensure
  Rationale dataset $R$ which contains Semantic Rationale (SR) and Impression Rationale (IR)
  \State Design $\&$ refine prompt pool for SR (SRP) and IR (IRP)
  \For{each sample ($V_i$, $S_i$, $T_i$, $Y_i$) in $M$}
    \State //Randomly select a prompt from SRP for SR
    \State $SR_{prompt}$ ← PromptPoolforSR($V_i$, $S_i$, $T_i$, $Y_i$)
    \State //Randomly select a prompt from IRP for IR
    \State $IR_{prompt}$ ← PromptPoolforIR($V_i$, $S_i$, $T_i$, $Y_i$)
    \State Produce SR and IR via GPT-4o
    \State $SR_i$ ← GPT-4o($V_i$, $S_i$, $T_i$, $Y_i$, $SR_{prompt}$)
    \State $IR_i$ ← GPT-4o($V_i$, $S_i$, $T_i$, $Y_i$, $IR_{prompt}$)
    \State Add ($V_i$, $S_i$, $T_i$, $Y_i$, $SR_i$, $IR_i$) to $R$
\EndFor
\end{algorithmic}
\end{algorithm}

To comprehensively capture the emotional rationale underlying the identified sentiment polarity from a semantic perspective of both image and text, we employ GPT-4o (gpt-4o-2024-05-13) via the OpenAI API\footnote{https://platform.openai.com} to generate SR. Meanwhile, to enable the model to effectively capture implicit emotional cues arising from the affective resonance of aesthetic attributes, we employ GPT-4o to generate the IR. 

To enhance the diversity of generated semantic and impression rationales (SR and IR), we designed and refined a series of templates to construct separate prompt pools for SR and IR, from which a prompt is randomly selected as instructions to guide GPT-4o in generating the corresponding rationale.
In this study, we adopt the approach outlined by Sarah~\etal~\cite{wiegreffe2022reframing} and Wang~\etal~\cite{wang2023scott}, leveraging tailored prompts conditioned on the dataset's gold-standard annotations to generate SR and IR using GPT-4o.
The example prompts for generating SR and IR are presented in Tables~\ref{tab:SRprompts} and~\ref{tab:IRprompts}, respectively. 
\begin{table}[t]
\centering
\caption{Example prompts for semantic rationale generation.}
\small 
\setlength{\tabcolsep}{4pt} 
\renewcommand{\arraystretch}{1.2} 
\begin{tabular}{@{}p{0.2\columnwidth}p{0.7\columnwidth}@{}}
\toprule
Type & Prompts \\ \midrule
System Prompt & 
You are an AI assistant specializing in multimodal understanding and sentiment analysis, particularly in scenarios involving the integration of image and text modalities. \\
\midrule
\multirow{3}{=}{Semantic Rationale Generation Prompt} & 
You will be provided with an image-text pair. Your task is to analyze the sentiment towards the specified entity \{aspect\} and explain why the sentiment polarity \{label\} is appropriate. \\
& 
Your explanation should consider both the semantic meaning of the text and the visual representation of the image, focusing on explicit content and the emotional or contextual cues conveyed by their combination. \\
& 
Start your response with: "Based on the image-text pair, the sentiment towards \{aspect\} is \{label\} because...". Provide a concise, focused explanation highlighting the single most compelling reason for this sentiment classification. \\
\bottomrule
\end{tabular}

\label{tab:SRprompts}
\end{table}

\begin{table}[t]
\centering
\caption{Example prompts for impression rationale generation.}
\small 
\setlength{\tabcolsep}{4pt} 
\renewcommand{\arraystretch}{1.2} 
\begin{tabular}{@{}p{0.2\columnwidth}p{0.7\columnwidth}@{}}
\toprule
Type & Prompts \\ \midrule
System Prompt & 
You are an AI assistant specializing in multimodal emotion and aesthetic understanding, especially in analyzing the emotional responses elicited by visual content. \\
\midrule
\multirow{3}{=}{Impression Rationale Generation Prompt} & 
You will be given an image-text pair. Your task is to analyze the specified entity \{aspect\} and its associated sentiment label \{label\} based entirely on the image's aesthetic attributes and the emotional resonance it conveys.\\
& 
Focus exclusively on the overall impression and visual connotations conveyed by the image, emphasizing why the assigned sentiment \{label\} aligns with the general mood or perception evoked by the entity. Avoid mentioning specific details; instead, highlight the prevailing emotional or aesthetic impression.
\\
\bottomrule
\end{tabular}

\label{tab:IRprompts}
\end{table}

\subsection{Linguistic-aware Semantic Alignment(LSA)}
\label{sec:LSA}
We first introduce dynamic patch selection in Sec.~\ref{sec:LSA-DPS}. Then, we introduce the semantic patch calibration in Sec.~\ref{sec:LSA-SPC}. and patch-token alignment in Sec.~\ref{sec:PTA}.
The overall process of LSA is shown in the persucode~\ref{alg:lsa}.
\subsubsection{Dynamic Patch Selection(DPS)}
\label{sec:LSA-DPS}

Dynamic Patch Selection (DPS) is considered a discriminative task that assigns significance scores to visual patches and selects valuable patches based on high scores. For the image in an image-text pair, we opt for vision Transformers as the visual encoder. The image $V$ is divided into $N_v$ non-overlapping patches by spatial distribution. 

These patches are then input as a visual token sequence into the vision Transformer to obtain a set of visual patch features $V=\left(v_{cls}, v_1, v_2, \ldots, v_{N_v}\right) \in \mathbb{R}^{(N_v + 1) \times d}$. For sentence $S$, a pre-trained Transformer serves as the textual encoder. The sentence is tokenized into $N_s$ tokens and processed by the encoder to extract linguistic features $S=\left(s_1, s_2, \ldots, s_{N_s}\right) \in \mathbb{R}^{N_s \times d}$. Subsequently, we incorporate spatial information from images into visual patch features and use an MLP-based score-sensitive prediction mechanism to learn significant scores:
\begin{equation}
p_i^s=\operatorname{\textit{Sigmoid}}\left(\mathbf{M L P}\left(\boldsymbol{v}_i\right)\right), i \in\{1,2, \ldots, N_v\},
\end{equation}
\REO{where $p_i^s \in [0,1]$ represents the importance score assigned to each visual patch. Moreover, achieving refined cross-modal alignment requires more than depending solely on a scoring mechanism to identify valuable visual patches without linguistic supervision~\cite{meng2022adavit, fu2024linguistic}. Consequently, we introduce linguistic context by calculating attentive scores between visual patches and the input sentence. First, we derive linguistic-aware scores $p_i^l$ through cross-attention between visual patches and linguistic features. Then, we enhance key visual content by computing self-attention within patches, producing image-prominent scores $p_i^e$:}
\begin{equation}
p_i^l=Norm\left(\boldsymbol{v}_i \cdot S / d\right), p_i^e=Norm\left(\boldsymbol{v}_i \cdot V / d\right),
\end{equation}
where $Norm(\cdot)$ denotes the normalization of scores to a range from 0 to 1. $S$ and $V$ represent the global embeddings for linguistic features and visual patches, respectively. These scores are integrated to derive the final value score:
\REO{
\begin{equation}
p_i^f=(1-\beta) p_i^s+\frac{\beta}{2}\left(p_i^l+p_i^e\right),
\end{equation}
}
where $\beta$ refers to the weight parameter. 
After obtaining the value score $p^f=\left(p_1^f, p_2^f, p_3^f, \ldots, p_{N_v}^f\right) \in \mathbb{R}^{N_v}$, we convert it into a binary decision matrix $\{0, 1\}^{N_v}$ to determine patch selection. This matrix is constructed using the Gumbel-Softmax technique~\cite{maddison2017concrete}, ensuring a smooth and differentiable sampling process. The Gumbel-Softmax matrix is defined as:
\begin{equation}
\boldsymbol{M}_{i, l}=\frac{\exp \left(\log \left(\boldsymbol{m}_{i, l}+G_{i, l}\right) / \tau\right)}{\sum_{j=1}^L \exp \left(\log \left(\boldsymbol{m}_{i, j}+G_{i, j}\right) / \tau\right)},
\end{equation}
where $M \in \mathbb{R}^{N_v \times L}$, $L$ indicates the total number of categories. \REO{In this scenario, $L$ is set to 2 for the binary decision ($\boldsymbol{m}_{i, 1}=p_i^f$, $\boldsymbol{m}_{i, 2}=1 - p_i^f$)}. $G_i=-\log \left(-\log \left(U_i\right)\right)$ represents the Gumbel distribution, $U_i$ refers to the uniform distribution and $\tau$ is the temperature parameter. 

Next, we obtain the differentiable decision matrix $D$ by applying the arg-max on $M$:
\begin{equation}
\boldsymbol{D}=\operatorname{Sampling}(\boldsymbol{M})_{*, 1} \in\{0,1\}^{N_v},
\end{equation}
where $D$ indicates patch selection outcomes: ``1'' for important patches and ``0'' for redundant ones. In the training stage, gradients are backpropagated through the differentiable decision matrix, enabling the dynamic selection of valuable visual patches via the score-sensitive prediction mechanism.

\begin{algorithm}[t]
\caption{\REO{Linguistic-aware Semantic Alignment (LSA)}}
\label{alg:lsa}
\begin{algorithmic}[1]
\Procedure{Dynamic Patch Selection}{$V, S$}
    \State Extract visual patches $V \gets \text{ViT}(V)$, text tokens $S \gets \text{TextEnc}(S)$
    \State Compute significance scores: $p^s_i \gets \text{MLP}(v_i)$, $p^l_i \gets \text{Norm}(v_i S^\top)$, $p^e_i \gets \text{Norm}(v_i V^\top)$
    \State Fuse scores: $p^f_i \gets (1-\beta)p^s_i + \frac{\beta}{2}(p^l_i + p^e_i)$
    \State Apply Gumbel-Softmax sampling to obtain binary mask $D \in \{0,1\}^{N_v}$
    \State Return selected patches $V^p \gets \{v_i | D_i=1\}$
\EndProcedure

\Procedure{Semantic Patch Calibration}{$V^p$}
    \State Aggregate key patches: $\tilde{V}^p \gets \text{Softmax}(\text{MLP}(V^p)) \cdot V^p$ \Comment{Adaptive weighting}
    \State Fuse redundant patches: $\tilde{v}^r \gets \sum \tilde{p}_i v_i$ \Comment{Weighted sum via $p^f$}
    \State Return $\tilde{V}^p \gets [v_{cls}; \tilde{V}^p; \tilde{v}^r]$ 
\EndProcedure

\Procedure{Patch-Token Alignment}{$\tilde{V}^p, S$}
    \State Compute cosine similarity matrix $A \in \mathbb{R}^{(N_f+2) \times N_s}$
    \State Calculate alignment score $K(V,S) \gets \frac{1}{2}\left(\text{mean}(\max_j A_{ij}) + \text{mean}(\max_i A_{ij})\right)$
    \State Optimize with $\mathcal{L}_{\text{align}} \gets \text{Bi-directional Triplet Loss}(K(V,S), K(V,\hat{S}), K(\hat{V},S))$
\EndProcedure
\end{algorithmic}
\end{algorithm}

\subsubsection{Semantic Patch Calibration(SPC)}
\label{sec:LSA-SPC}
This section aims to further refine the semantic representation of the selected valuable visual patches.
After dynamically selecting important visual patches guided by linguistic supervision, we designate them as $V^p=\left(v^p_1, v^p_2, \ldots, v^p_{N_p}\right) \in \mathbb{R}^{N_p \times d}$. $N_p$ is the number of selected valuable visual patches. We employ an aggregation network~\cite{zong2022self} to model multiple aggregation weights and combine the selected $N_p$ visual patches to generate $N_f$ informative visual features:
\begin{equation}
\tilde{\boldsymbol{v}}_j^p=\sum_{i=1}^{N_p}(\boldsymbol{W})_{i j} \cdot \boldsymbol{v}^p_i, \quad j=\left[1, \ldots, N_f\right],
\end{equation}
\begin{equation}
\boldsymbol{W}=\operatorname{\textit{softmax}}\left(\mathbf{M L P}\left(\boldsymbol V^p \right)\right),
\end{equation}
where $(\boldsymbol{W})$ denotes the normalized weight matrix and $\sum_{i=1}^{N_s}(\boldsymbol{W})_{i j}=1$. $N_f$ is the number of aggregated patches
($N_f < N_p$). The aggregation network adaptively combines visually similar patches and is differentiable for end-to-end training. While redundant visual patches can be discarded, they may contain supplementary semantic features for refined cross-modal alignment. Therefore, we fuse them into a single patch:
\begin{equation}
\tilde{\boldsymbol{v}}^r=\sum_{i \in \mathcal{N}} \tilde{p}_i \cdot \boldsymbol{v}_i, \quad \tilde{p}_i=\frac{\exp \left(p_i^f\right) \boldsymbol{D}_i}{\sum_{i=1}^N \exp \left(p_i^f\right) \boldsymbol{D}_i},
\end{equation}
where $\mathcal{N}$ represents the set for redundant visual patches. $\tilde{p}_i$ denotes the normalized score of the value score $p_i^f$. Finally, this component models the calibrated refined visual patches, denoted as $\tilde{V}^p=\left(v_{cls}, \tilde{v}^p_1, \tilde{v}^p_2, \ldots, \tilde{v}^p_{N_f}, \tilde{v}^r \right) \in \mathbb{R}^{(N_f + 2) \times d}$.

\subsubsection{Patch-token Alignment(PTA)}
\label{sec:PTA}

This module aims to achieve the fine-grained patch-token level alignment. Specifically, we first utilize the refined visual patches $\tilde{V}^p$ and linguistic features $S$ to compute token-wise similarities, producing a patch-token similarity matrix $A \in \mathbb{R}^{(N_f + 2) \times N_s}$. $(A)_{i j}=\frac{\left(\tilde{\boldsymbol{v}}_i\right)^T \boldsymbol{s}_j}{\left\|\tilde{\boldsymbol{v}}_i\right\|\left\|\boldsymbol{s}_j\right\|}$ denotes the patch-token level alignment score between the $i$-th visual patch and the $j$-th word. Subsequently, maximum-correspondence interaction is introduced to aggregate cross-modal alignment. For each visual patch (or token), we identify the most aligned textual token (or patch) and calculate the average alignment score $K(V,S)$, representing the overall alignment between the image $V$ and the sentence $S$:
\begin{equation}
K(V, S)=\frac{1}{N_f + 2} \sum_{i=1}^{N_f + 2} \max _j(\boldsymbol{A})_{i j}+\frac{1}{N_s} \sum_{j=1}^{N_s} \max _i(\boldsymbol{A})_{i j}
\end{equation}

Following a previous method~\cite{faghri2017vse++}, the bi-direction triplet loss with hard negative mining is exploited:
\begin{equation}
\begin{aligned}
\mathcal{L}_{\text {align }}= & \sum_{(V, S)}[\gamma-K(V, S)+K(V, \hat{S})]_{+} \\
& +[\gamma-K(V, S)+K(\hat{V}, S)]_{+},
\end{aligned}
\end{equation}
where $\gamma$ is the trade-off parameter. $[x]_{+}=\max (x, 0)$ and $(V, S)$ refers to a positive image-text pair in the mini-batch. Moreover, $\hat{S}=\operatorname{argmax}_{j \neq \boldsymbol{S}} K(V, j)$ and $\hat{V}=\operatorname{argmax}_{i \neq \boldsymbol{V}} K(i, V)$ indicate the hardest negative sentence and visual examples within a mini-batch, respectively.

\subsection{Rationale-aware Learning}

To endow the model with the ability to perform semantic causality and impression reasoning, we propose a rationale-aware learning framework designed to fine-tune a sequence-to-sequence (seq2seq) model. 
This seq2seq model is proposed to achieve three task objectives for each specific target within the image-text pair: sentiment classification (SC), semantic rationale generation (SRG), and impression rationale generation (IRG). 
These tasks are differentiated by the use of distinct input configurations and input content. For SC, the decoder outputs only the predicted sentiment polarity. In SRG and IRG, the decoder produces the corresponding rationale and the sentiment prediction.

Specifically, our input comprises the textual sentence $S=\left(s_1, s_2, \ldots, s_{N_s}\right)$, the overall aesthetic caption of the image $A^c=\left(a^c_1, a^c_2, \ldots, a^c_{N_c}\right)$, the object-level description $A^o=\left(a^o_1, a^o_2, \ldots, a^o_{N_o}\right)$, which pertains to either facial or aesthetic attributes and the specific target $T$. The input format is determined by the presence of the specific target within the visual content. For example, if the specific target is identified in the image, based on the annotations provided by Wang~\etal~\cite{wang2023fine}, the input for SC, SRG, and IRG is defined as follows:
\begin{equation}
H^{\mathrm{sc}}=encoder(t_{\langle\mathrm{sc}\rangle}, A^c, S, T),
\end{equation}
\begin{equation}
H^{\mathrm{srg}}=encoder(t_{\langle\mathrm{srg}\rangle}, A^c, S, T),
\end{equation}
\begin{equation}
H^{\mathrm{irg}}=encoder(t_{\langle\mathrm{irg}\rangle}, A^c, S, T),
\end{equation}
where $encoder(\cdot)$ is the Transformer encoder of the seq2seq model. The tokens $t_{\langle\mathrm{sc}\rangle}$, $t_{\langle\mathrm{srg}\rangle}$, and $t_{\langle\mathrm{irg}\rangle}$ are specialized tokens designed to represent distinct tasks. 
Although the specific aspects are not present in the image, this does not imply that sentimental cues from the image have no impact on predicting the sentiment polarity. On the contrary, incorporating sentiment cues from the holistic image can provide valuable insights into the influence of image aesthetic attributes on the sentiment prediction for the specific aspect.
For samples where specific targets are present in the visual content, the input format is structured as follows: 
\begin{equation}
H^{\mathrm{sc}}=encoder(t_{\langle\mathrm{sc}\rangle}, S, A^o, T),
\end{equation}
\begin{equation}
H^{\mathrm{srg}}=encoder(t_{\langle\mathrm{srg}\rangle}, S, A^o, T),
\end{equation}
\begin{equation}
H^{\mathrm{irg}}=encoder(t_{\langle\mathrm{irg}\rangle},S, A^o, T).
\end{equation}

We employ fine-grained, object-level emotion-laden descriptions to establish alignment between specific targets and their corresponding objects in the image, which enhances both the accuracy and interpretability of the sentiment prediction process.
Subsequently, these hidden features are passed through a stack of self-attention-based encoders, which dynamically fuse representations and model both intra-modal and cross-modal interactions.
Finally, the decoder produces task-specific outputs. For Sentiment Classification (SC), the decoder generates the predicted sentiment polarity, selecting from ``positive,'' ``negative,'' or ``neutral,'' denoted as $\hat{y}_{sc}$:
\begin{equation}
G^{\mathrm{sc}}=\left[\langle\operatorname{sen}\rangle \hat{y}^{\mathrm{sc}}\langle/ \operatorname{sen}\rangle\right],
\end{equation}
where the special tokens $\langle\operatorname{sen}\rangle$ and $\langle/ \operatorname{sen}\rangle$ are denoted as the start and end markers for SC predictors.
For the two additional rationale generation tasks SRG and IRG, the decoder generates not only the semantic rationale $\hat{sr}$ and impression rationale $\hat{ir}$ for the specific target but also their corresponding sentiment predictions $\hat{y}_{sr}$ and $\hat{y}_{si}$:
\begin{equation}
G^{\mathrm{sr}}=\left[\langle\mathrm{sr}\rangle \hat{sr}\langle/ \mathrm{sr}\rangle\langle\operatorname{sen}\rangle \hat{y}^{\mathrm{sr}}\langle/ \operatorname{sen}\rangle\right],
\end{equation}
\begin{equation}
G^{\mathrm{ir}}=\left[\langle\mathrm{ir}\rangle \hat{ir}\langle/ \mathrm{ir}\rangle\langle\operatorname{sen}\rangle \hat{y}^{\mathrm{ir}}\langle/ \operatorname{sen}\rangle\right],
\end{equation}
where $\langle\mathrm{sr}\rangle$, $\langle/ \mathrm{sr}\rangle$, $\langle\mathrm{ir}\rangle$, $\langle/ \mathrm{ir}\rangle$, $\langle\mathrm{sen}\rangle$, and $\langle/ \mathrm{sen}\rangle$ serve as specialized markers to delineate the rationale and sentiment polarity.
Finally, the input sequence is uniformly denoted as $X$, and the generated textual content is represented as $Z=\left\{z_1, z_2, \ldots, z_{N_z}\right\}$. Consequently, the loss function for the generation process is formulated as follows:
\begin{equation}
\label{eq:gene}
\mathcal{L}_Z=-\frac{1}{N} \sum_{i=1}^N \sum_{n_z=1}^{N_z} \log P\left(z_{i, n_z} \mid \hat{z}_{i,<n_z}, X\right),
\end{equation}
where $z_{i,n_z}$ is the ground truth token at position $n_z$ for sample $i$, $\hat{z}_{i,<n_z}$ represents the generated sequence up to position $n_z-1$ for sample $i$, and $P(z_{i,n_z} \mid \hat{z}_{i,<n_z}, X)$ denotes the probability of generating token $z_{i,n_z}$ conditioned on $\hat{z}_{i,<n_z}$ and $X$. \REO{In this rationale-aware learning framework, since all objectives are formulated as generative tasks, the loss functions $\mathcal{L}_{SC}$, $\mathcal{L}_{SRG}$, and $\mathcal{L}_{IRG}$ are all employ the generative loss function, E.q.~\ref{eq:gene}.
Therefore, the objective function in the proposed method is formulated as follows:}
\begin{equation}
\mathcal{L}=\alpha \mathcal{L}_{\mathrm{SC}}+\frac{1-\alpha}{2} \mathcal{L}_{\mathrm{SRG}}+\frac{1-\alpha}{2} \mathcal{L}_{\mathrm{IRG}} + \lambda \mathcal{L}_{align},
\end{equation}
where $\alpha, \lambda \in (0, 1)$ are tradeoff hyperparameters that regulate the relative contributions of each generative loss and the patch-token alignment.

\begin{table*}[!thb]
\centering
\caption{Detailed Statistics of Twitter-2015, Twitter-2017, and Political Twitter datasets. The "\#sentence" refers to the total number of sentences. "\#Avg. Length" denotes the average length of sentences, while "\#Avg. Aspect" indicates the average number of aspects in a sentence. "\#Avg. Length of SR", "\#Avg. Length of IR", "\#Avg. Length of AC", "\#Avg. Length of FD", and "\#Avg. Length of AO" correspond to the average lengths of semantic rationales (SR), impression rationales (IR), aesthetic captions for the entire image, facial descriptions, and aesthetic captions for objects.}
\begin{adjustbox}{width=0.8\textwidth}
\begin{tabular}{@{}lccccccccc@{}}
\toprule
\multirow{2}{*}{Label} & \multicolumn{3}{c}{Twitter-2015} & \multicolumn{3}{c}{Twitter-2017} & \multicolumn{3}{c}{Political Twitter} \\ \cmidrule(l){2-10} 
            & Train   & Dev   & Test   & Train   & Dev   & Test   & Train    & Dev    & Test    \\ \midrule
Positive        & 928    & 303   & 317   & 1508   & 515   & 493   & 3318    & 570    & 176    \\
Neutral        & 1883   & 670   & 607   & 1638   & 517   & 573   & 4697    & 823    & 368    \\
Negative        & 368    & 149   & 113   & 416    & 144   & 168   & 887     & 166    & 305    \\
Total         & 3179   & 1122   & 1037   & 3562   & 1176   & 1234   & 8902    & 1559    & 849    \\ \hdashline
\#Sentence       & 2101   & 727   & 674   & 1746   & 577   & 587   & 5105    & 900    & 407    \\
\#Avg. Length     & 16.72   & 16.74  & 17.05  & 16.21   & 16.37  & 16.38  & 16.62    & 16.67   & 16.59   \\
\#Avg. Aspect     & 1.51   & 1.54   & 1.54   & 2.04   & 2.04   & 2.10   & 1.74    & 1.73    & 2.09    \\ \midrule
\#Avg. Length of SR  & 42.5   & 42.4   & 42.5   & 42.6   & 42.8   & 43.0   & 42.7    & 42.6    & 42.2    \\
\#Avg. Length of IR  & 56.7   & 56.0   & 55.7   & 55.5   & 56.1   & 55.4   & 55.9    & 56.1    & 56.3    \\
\#Avg. Length of AC  & 35.9   & 35.9   & 35.5   & 32.5   & 32.5   & 31.6   & 34.0    & 34.2    & 33.3    \\
\#Avg. Length of FD  & 39.2   & 38.5   & 37.8   & 38.9   & 38.5   & 39.3   & 39.0    & 38.4    & 38.7    \\
\#Avg. Length of AO  & 29.1   & 29.7   & 30.3   & 28.9   & 29.4   & 28.9   & 29.1    & 29.1    & 31.3    \\ \bottomrule
\end{tabular}
\end{adjustbox}
\label{tab:twitter}
\end{table*}

\section{EXPERIMENTS}
\label{sec:result}
In this section, we provide a comprehensive description of the experimental settings and evaluate the proposed method on three publicly available MASC datasets, benchmarking it against state-of-the-art methods. Furthermore, we perform an extensive series of studies to thoroughly analyze the effectiveness of the proposed approach.

\subsection{Experimental Settings}

\subsubsection{Datasets}
We utilize three widely recognized benchmark datasets for MASC~\cite{yuadapting, yang2021fine}: Twitter-2015, Twitter-2017, and the Political Twitter dataset. Each sample within these datasets comprises a user-generated multimodal image-text pair, including an image, a textual sentence, and one or more specific targets. Each aspect is annotated with a sentiment label from the set {Positive, Negative, Neutral}. The detailed statistics of these datasets are presented in Table~\ref{tab:twitter}. Furthermore, we incorporate semantic rationale (SR), impression rationale (IR), aesthetic captions for the entire image (AC), facial descriptions (FD), and aesthetic captions for objects (AO) for each data point. The maximum length for facial descriptions and aesthetic captions is constrained to 50 tokens.
\subsubsection{Implementation Details}
We adopt the seq2seq model Flan-T5~\cite{chung2024scaling} as the backbone of our generative framework. Specifically, the model is trained for 10 epochs using the AdamW optimizer~\cite{loshchilov2017decoupled}, with a batch size of 4. A grid search is performed on the development set to determine the optimal learning rate, $\alpha$ and $\lambda$ for Flan-T5 across the three datasets. \REO{The selected values for learning rate are $3e-4$, $3e-4$, $1e-4$, respectively, for the Twitter-2015, Twitter-2017 and Political Twitter.} \REO{The trade-off hyper-parameter sets ($\alpha$ and $\lambda$) are 0.2, 0.1, 0.2 and 0.2, 0.5, 0.5, respectively, for the Twitter-2015, Twitter-2017 and Political Twitter.} Consistent with prior research on MASC~\cite{yuadapting, fan2024dual}, we employ Accuracy (Acc) and F1 score (F1) as the evaluation metrics. The model is implemented using PyTorch, and experiments are conducted on an NVIDIA V100 GPU with 30 GB of memory.

\subsection{Compared Baselines}

We conducted a comprehensive comparative evaluation of the proposed method against a range of robust baseline approaches, which are classified into three categories. The first category consists of image-only methods:
\begin{itemize}
\item \textbf{Res-Target}~\cite{he2016deep} leverages ResNet as its backbone to extract visual features exclusively for predicting the sentiment of the specified target.

\end{itemize}
The second category includes text-only approaches:
\begin{itemize}
\item \textbf{MemNet}~\cite{tang2016aspect} employs a stacked architecture of multiple memory networks to build deep memory networks.

\item \textbf{MGAN}~\cite{fan2018multi} is based on a multi-grained attention architecture designed to adaptively capture both coarse-grained and fine-grained interactions.

\item \textbf{BERT}~\cite{kenton2019bert} is a powerful pre-trained language model trained using a masked language modeling objective and next sentence prediction.

\end{itemize}

Finally, this study incorporates the following advanced image-text multimodal approaches:
\begin{itemize}
\item \textbf{MIMN}~\cite{xu2019multi} comprises two customized interactive memory networks designed to capture both inter-modal dynamics between different modalities and intra-modal dynamics within each individual modality.


\item \textbf{ESAFN}~\cite{yu2019entity} is a target-sensitive interaction and fusion network designed to adaptively capture interactive features across modalities while also modeling intra-modality features.

\item \textbf{TomBERT}~\cite{yuadapting} utilizes BERT and ResNet as backbone models for encoding textual and visual content, respectively. Cross-modal fusion is accomplished by integrating these features into a BERT encoder.

\item \textbf{JML-MASC}~\cite{ju2021joint} jointly extracts the specific targets and identifies their sentiment polarity by utilizing a visual de-nosing mechanism and attention-based fusion framework.

\item \textbf{EF-CapTrBERT}~\cite{khan2021exploiting} converts visual content into an auxiliary sentence, which is then combined with the input sentence and processed through a BERT encoder for sentiment prediction.

\item \textbf{VLP-MABSA}~\cite{ling2022vision} is a task-specific pre-trained generative framework for multimodal aspect-based sentiment analysis, built on the BART architecture. 

\item \textbf{FITE}~\cite{yang2022face} is a translation-based approach, which captures facial features in the image and translates them into a corresponding facial description as an auxiliary sentence for sentiment classification.

\item \textbf{CMMT-MASC}~\cite{yang2022cross} is a cross-modal multi-task Transformer designed for MASC. Additionally, it employs multimodal gating mechanisms to dynamically regulate the flow of textual and visual information during interactions.

\item \textbf{HIMT}~\cite{yu2022hierarchical} is a Transformer framework that incorporates a hierarchical interaction component to model the relationships between specific aspects and the input sentence, as well as the interactions between specific aspects and object-level visual content.

\item \textbf{IMT}~\cite{yu2022targeted} is a coarse-to-fine-grained multimodal matching network that predicts image-target relevance and performs object-target alignment to support sentiment polarity identification.

\item \textbf{CoolNet}~\cite{xiao2023cross} is a fine-grained cross-modal alignment approach that aligns textual and visual content from both semantic and syntactic perspectives. 

\item \textbf{UnifiedTMSC}~\cite{liu2023descriptive} introduces a descriptive prompt paraphrasing paradigm to generate paraphrased prompts, while optimizing image vectors within the multimodal representation space of vision and language.

\item \textbf{VEMP}~\cite{yang2023visual} decodes the semantic information of visual elements by utilizing textual tokens in the image, target-aware adjective-noun pairs, and image captions. 

\item \textbf{Atlantis-MASC}~\cite{xiao2024atlantis} is a trident-shaped, aesthetics-driven approach for joint MABSA, which integrates image aesthetic attributes and achieves effective alignment of vision and text across multiple granular levels.

\item \textbf{MDCA}~\cite{fan2024dual} is a generative framework proposed to provide supplementary reasoning and explicit rationales to explain why specific content conveys certain sentiment.
\end{itemize}

\begin{table*}[]
\centering
\caption{The main results ($\%$) are presented with the best-performing results highlighted in \textbf{bold} and the second-best values indicated with \underline{underlined text}.}
\begin{tabular}{@{}lllcccccc@{}}

\toprule
\multirow{2}{*}{Modality}        & \multirow{2}{*}{Model} & \multirow{2}{*}{Venue} & \multicolumn{2}{c}{Twitter-2015} & \multicolumn{2}{c}{Twitter-2017} & \multicolumn{2}{c}{\REO{Political Twitter}} \\ \cmidrule(l){4-9} 
                                 &                        &                        & Acc             & F1       & Acc             & F1       & \REO{Acc}               & \REO{F1}          \\ \midrule
Image Only                       & Res-Target             & CVPR 2016              & 59.88           & 46.48          & 58.59           & 53.98          & \REO{60.21}             & \REO{58.42}             \\ \midrule
\multirow{3}{*}{Text Only}       & MemNet                 & EMNLP 2016             & 70.11           & 61.76          & 64.18           & 60.90          & -                 & -                 \\
                                 & MGAN                   & EMNLP 2018             & 71.17           & 64.21          & 64.75           & 61.46          & \REO{67.37}             & \REO{62.78}             \\
                                 & BERT                   & NAACL 2019             & 74.15           & 68.86          & 68.15           & 65.23          & \REO{69.41}             & \REO{64.25}             \\ \midrule
\multirow{15}{*}{Image and Text} & MIMN                   & AAAI 2019              & 71.84           & 65.69          & 65.88           & 62.99          & \REO{70.52}             & \REO{65.39}             \\
                                 & ESAFN                  & TASLP 2019             & 73.38           & 67.37          & 67.83           & 64.22          & \REO{69.22}             & \REO{64.66}             \\
                                 & TomBERT                & IJCAI 2019             & 77.15           & 71.15          & 70.34           & 68.03          & \REO{69.65}             & \REO{62.35}             \\
                                 & JML-MASC               & EMNLP 2021             & 78.70           & -              & 72.70           & -              & \REO{70.14}             & \REO{68.37}             \\
                                 & EF-CapTrBERT           & ACM MM 2021            & 78.01           & 73.25          & 69.77           & 68.42          & \REO{69.04}             & \REO{64.94}             \\
                                 & VLP-MABSA              & ACL 2022               & 78.60           & 73.80          & 73.80           & 71.80          & \REO{70.32}             & \REO{69.64}             \\
                                 & CMMT-MASC              & IPM 2022               & 77.90           & -              & 73.8            & -              & -                 & -                 \\
                                 & FITE                   & EMNLP 2022             & 78.49           & 73.90          & 70.90           & 68.70          & \REO{68.64}             & \REO{65.83}             \\
                                 & HIMT                   & TAC 2022               & 78.14           & 73.68          & 71.14           & 69.16          & -                 & -                 \\
                                 & IMT                    & IJCAI 2022             & 78.27           & 74.19          & 72.61           & 71.97          & \REO{69.92}             & \REO{67.86}             \\
                                 & CoolNet                & IPM 2023               & 79.92           & 75.28          & 71.64           & 69.58          & \REO{70.91}             & \REO{70.25}             \\
                                 & UnifiedTMSC            & EMNLP 2023             & 79.80           & 76.30          & \underline{75.40}           & \textbf{74.70} & -                 & -                 \\
                                 & VEMP                   & EMNLP 2023             & 78.88           & 75.09          & 73.01           & 72.42          & -                 & -                 \\
                                 & Atlantis-MASC          & INFFUS 2024            & 79.03           & -              & 74.20           & -              & \REO{69.83}             & \REO{68.97}             \\
                                 & MDCA                   & TNNLS 2024             & \underline{80.71}           & \underline{77.15}          & 73.91           & 72.37          & \REO{\underline{71.38}}             & \REO{\underline{70.94}}             \\ \midrule
Ours                             & Chimera                & -                      & \textbf{81.61}  & \textbf{77.98} & \textbf{75.62}  & \underline{74.59}          & \REO{\textbf{72.56}}    & \REO{\textbf{72.32}}    \\ \bottomrule
\end{tabular}
\label{exp:main_result}
\end{table*}

\subsection{Main Results}
The main results are presented in Table~\ref{exp:main_result}. Given that the two additional rationale generation tasks contribute to improving sentiment prediction by providing explanations for the underlying causes of sentiment, we select the prediction results from sentiment classification $\hat{y}^{\mathrm{sc}}$ as the primary outcomes for accuracy and F1 score evaluation. As presented in Table~\ref{exp:main_result}, the proposed method demonstrates competitive performance on both Twitter datasets compared to strong baselines from both text-only and multimodal approaches. Specifically, it achieves the highest accuracy (81.61$\%$) and F1 score (77.98$\%$) on the Twitter-2015 dataset, as well as the best accuracy (75.62$\%$) and a near-optimal F1 score (74.59$\%$) on the Twitter-2017 dataset.

Compared to the image-only approach (Res-Target), the proposed method achieves a remarkable improvement of over 21.73$\%$ in accuracy on the Twitter-2015 dataset. Similarly, when compared to the best-performing text-only method (BERT), the proposed method demonstrates a substantial performance gain, with a 7.46$\%$ increase in accuracy and a 9.12$\%$ improvement in F1 on Twitter-2015. These observations underscore the limitations of single-modality approaches in capturing subtle sentiment cues from multimodal content. Moreover, the proposed method consistently outperforms recent multimodal models, such as UnifiedTMSC, Atlantis-MASC, and MDCA. For instance, UnifiedTMSC adopts a paraphrasing-based approach to enrich textual features but lacks explicit modeling of visual aesthetic-driven affective impact. On Twitter-2017, the proposed method achieves comparable F1 performance (74.59 vs. 74.70) while delivering higher accuracy (75.62 vs. 75.40), which highlights the complementary benefits of aesthetic affective resonance modeling. While Atlantis-MASC incorporates image aesthetics, it primarily relies on global alignment techniques, which may overlook the intricate relationships between aspects and objects. The proposed method surpasses Atlantis-MASC by 1.58$\%$ in accuracy on Twitter-2017, underscoring the efficacy of its patch-token level and object-level alignment in capturing aspect-specific visual details. While MDCA incorporates reasoning and direct causality to explain sentiment causes, it primarily emphasizes textual semantic reasoning, which restricts its ability to effectively capture detailed visual content and the corresponding aesthetic affective resonance. In contrast, the proposed method surpasses MDCA with a 0.90$\%$ improvement in accuracy and a 0.83$\%$ increase in F1 on the Twitter-2015 dataset. This performance gain highlights the advantages of comprehensively understanding sentiment causality from both visual-textual semantic and affective resonance perspectives.

\subsection{Results on Political Twitter}
\REO{
The Political Twitter dataset differs significantly from Twitter-2015 and Twitter-2017, especially due to its challenging domain shift between training, development, and test sets. Such domain differences create substantial barriers to generalization, which makes the task particularly suitable for advanced models that can comprehend subtle causality and context shifts.
}\\

\REO{
From Table~\ref{exp:main_result}, it can be observed that the proposed Chimera demonstrates distinct advantages over existing approaches on the Political Twitter dataset. Compared to the third best performing method CoolNet, which achieved 71.32$\%$ accuracy and 69.64$\%$ F1 score, Chimera showcases a significant improvement. Similarly, MDCA, which performed with an accuracy of 71.38$\%$ and an F1 score of 70.94$\%$, still lags behind Chimera.
Additionally, we observed that the discrepancy between accuracy and F1-score significantly narrows as accuracy increases, particularly when accuracy surpasses 70$\%$. We hypothesize that the underlying cause may lie in the relatively balanced class distribution of sentiment categories (e.g., positive, neutral, negative) within the Political Twitter test set (as shown in Table~\ref{tab:twitter}). At higher accuracy levels, the ratios of false positives to false negatives exhibit increasing symmetry across models. This equilibrium consequently reduces the divergence between precision and recall metrics, thereby causing the F1-score – defined as their harmonic mean – to naturally converge with accuracy.}


\begin{table*}
\caption{The results ($\%$) of the ablation study for our Chimera model are presented. The top-performing values emphasized in \textbf{bold} and the second-best values distinguished using \underline{underlined text}. The notations “w/o SRG,” “w/o IRG,” and “w/o SRG $\&$ IRG” denote the exclusion of the respective generative tasks. \REO{“w/o IRG $\&$ AC” refers to the removal of IR generation task and replace the aesthetic caption (AC) with general caption.} “w/o LSA” represents the removal of the Linguistic-aware Semantic Alignment branch, while “w/o OD” indicates the exclusion of object-level descriptions (e.g., facial descriptions and object-level aesthetic captions) from the input sequence.}
  \resizebox{\textwidth}{!}{

\begin{tabular}{@{}lcccccccccccccccccccccccccc@{}}
\toprule
\multirow{3}{*}{\textbf{Method}} & \multicolumn{8}{c}{\textbf{Twitter-2015}}                                                                                                                                                                           & \multicolumn{1}{l}{} & \multicolumn{8}{c}{\textbf{Twitter-2017}}                                                                                                                                                                           & \multicolumn{1}{l}{} & \multicolumn{8}{c}{\textbf{Political Twitter}}                                                                                                                                                                      \\ \cmidrule(l){2-27} 
                                 & \textbf{Acc}              & \textbf{F1}               &                      & \textbf{Acc}              & \textbf{F1}               &                      & \textbf{Acc}              & \textbf{F1}               &                      & \textbf{Acc}              & \textbf{F1}               &                      & \textbf{Acc}              & \textbf{F1}               &                      & \textbf{Acc}              & \textbf{F1}               &                      & \textbf{Acc}              & \textbf{F1}               &                      & \textbf{Acc}              & \textbf{F1}               &                      & \textbf{Acc}              & \textbf{F1}               \\ \cmidrule(lr){2-9} \cmidrule(lr){11-18} \cmidrule(l){20-27} 
                                 & \multicolumn{2}{c}{\textbf{SC}}                       &                      & \multicolumn{2}{c}{\textbf{SRG}}                      &                      & \multicolumn{2}{c}{\textbf{IRG}}                      &                      & \multicolumn{2}{c}{\textbf{SC}}                       &                      & \multicolumn{2}{c}{\textbf{SRG}}                      &                      & \multicolumn{2}{c}{\textbf{IRG}}                      &                      & \multicolumn{2}{c}{\textbf{SC}}                       &                      & \multicolumn{2}{c}{\textbf{SRG}}                      &                      & \multicolumn{2}{c}{\textbf{IRG}}                      \\ \cmidrule(r){1-3} \cmidrule(lr){5-6} \cmidrule(lr){8-9} \cmidrule(lr){11-12} \cmidrule(lr){14-15} \cmidrule(lr){17-18} \cmidrule(lr){20-21} \cmidrule(lr){23-24} \cmidrule(l){26-27} 
Chimera                          & \textbf{81.61}            & \textbf{77.98}            &                      & \underline{81.12}               & \underline{77.11}               &                      & \underline{77.56}               & \underline{73.55}               &                      & \textbf{75.62}            & \textbf{74.59}            &                      & \underline{75.09}               & \underline{73.64}               &                      & \underline{71.96}               & \underline{68.23}               &                      & \textbf{72.56}            & \textbf{72.32}            &                      & \underline{71.69}               & \underline{71.40}               &                      & \underline{69.30}               & \underline{68.95}               \\ \midrule
w/o SRG                          & 80.52                     & 76.10                     &                      & -                         & -                         &                      & 75.83                     & 70.96                     &                      & 73.50                     & 72.49                     &                      & -                         & -                         &                      & 70.66                     & 67.20                     &                      & 70.43                     & 69.88                     &                      & -                         & -                         &                      & 68.25                     & 67.58                     \\
w/o IRG                          & 80.23                     & 75.22                     &                      & 80.03                     & 75.42                     &                      & -                         & -                         &                      & 71.88                     & 70.16                     &                      & 72.6                      & 70.73                     &                      & -                         & -                         &                      & 71.15                     & 70.70                     &                      & 71.01                     & 70.52                     &                      & -                         & -                         \\
\REO{w/o IRG \& AC}                    & \REO{80.67}                     & \REO{76.03}                     &                      & \REO{80.11}                     & \REO{76.46}                     &                      & -                         & -                         & \multicolumn{1}{l}{} & \multicolumn{1}{l}{\REO{71.59}} & \multicolumn{1}{l}{\REO{69.83}} & \multicolumn{1}{l}{} & \REO{72.25}                     & \REO{70.33}                     & \multicolumn{1}{l}{} & -                         & -                         & \multicolumn{1}{l}{} & \multicolumn{1}{l}{\REO{70.62}} & \multicolumn{1}{l}{\REO{70.06}} & \multicolumn{1}{l}{} & \multicolumn{1}{l}{\REO{71.04}} & \multicolumn{1}{l}{\REO{70.47}} & \multicolumn{1}{l}{} & -                         & -                         \\
w/o SRG \& IRG                   & 77.24                     & 71.82                     &                      & -                         & -                         &                      & -                         & -                         &                      & 71.23                     & 68.98                     &                      & -                         & -                         &                      & -                         & -                         &                      & 67.88                     & 67.20                     &                      & -                         & -                         &                      & -                         & -                         \\
w/o LSA                          & \multicolumn{1}{l}{80.54} & \multicolumn{1}{l}{77.03} & \multicolumn{1}{l}{} & \multicolumn{1}{l}{79.75} & \multicolumn{1}{l}{76.22} & \multicolumn{1}{l}{} & \multicolumn{1}{l}{76.52} & \multicolumn{1}{l}{72.03} & \multicolumn{1}{l}{} & \multicolumn{1}{l}{73.72} & \multicolumn{1}{l}{70.96} & \multicolumn{1}{l}{} & \multicolumn{1}{l}{74.38} & \multicolumn{1}{l}{72.26} & \multicolumn{1}{l}{} & \multicolumn{1}{l}{71.36} & \multicolumn{1}{l}{67.88} & \multicolumn{1}{l}{} & \multicolumn{1}{l}{71.86} & \multicolumn{1}{l}{71.37} & \multicolumn{1}{l}{} & \multicolumn{1}{l}{70.92} & \multicolumn{1}{l}{70.55} & \multicolumn{1}{l}{} & \multicolumn{1}{l}{68.43} & \multicolumn{1}{l}{67.99} \\
w/o OD                           & \multicolumn{1}{l}{79.96} & \multicolumn{1}{l}{76.08} & \multicolumn{1}{l}{} & \multicolumn{1}{l}{80.09} & \multicolumn{1}{l}{76.32} & \multicolumn{1}{l}{} & \multicolumn{1}{l}{77.12} & \multicolumn{1}{l}{72.84} & \multicolumn{1}{l}{} & \multicolumn{1}{l}{73.06} & \multicolumn{1}{l}{70.85} & \multicolumn{1}{l}{} & \multicolumn{1}{l}{74.37} & \multicolumn{1}{l}{72.36} & \multicolumn{1}{l}{} & \multicolumn{1}{l}{71.11} & \multicolumn{1}{l}{67.53} & \multicolumn{1}{l}{} & \multicolumn{1}{l}{71.64} & \multicolumn{1}{l}{71.12} & \multicolumn{1}{l}{} & \multicolumn{1}{l}{71.12} & \multicolumn{1}{l}{70.77} & \multicolumn{1}{l}{} & \multicolumn{1}{l}{68.55} & \multicolumn{1}{l}{68.07} \\
w/o Aes-cap                      & \multicolumn{1}{l}{80.03} & \multicolumn{1}{l}{75.27} & \multicolumn{1}{l}{} & \multicolumn{1}{l}{79.94} & \multicolumn{1}{l}{76.05} & \multicolumn{1}{l}{} & \multicolumn{1}{l}{75.69} & \multicolumn{1}{l}{71.08} & \multicolumn{1}{l}{} & \multicolumn{1}{l}{72.36} & \multicolumn{1}{l}{71.64} & \multicolumn{1}{l}{} & \multicolumn{1}{l}{72.28} & \multicolumn{1}{l}{71.21} & \multicolumn{1}{l}{} & \multicolumn{1}{l}{69.28} & \multicolumn{1}{l}{65.44} & \multicolumn{1}{l}{} & \multicolumn{1}{l}{69.43} & \multicolumn{1}{l}{68.94} & \multicolumn{1}{l}{} & \multicolumn{1}{l}{69.37} & \multicolumn{1}{l}{69.00} & \multicolumn{1}{l}{} & \multicolumn{1}{l}{67.85} & \multicolumn{1}{l}{67.27} \\ \bottomrule
\end{tabular}
}
\label{exp:ablation}
\end{table*}

\begin{figure*}[t!]
	\centerline{\includegraphics[width=1\textwidth]{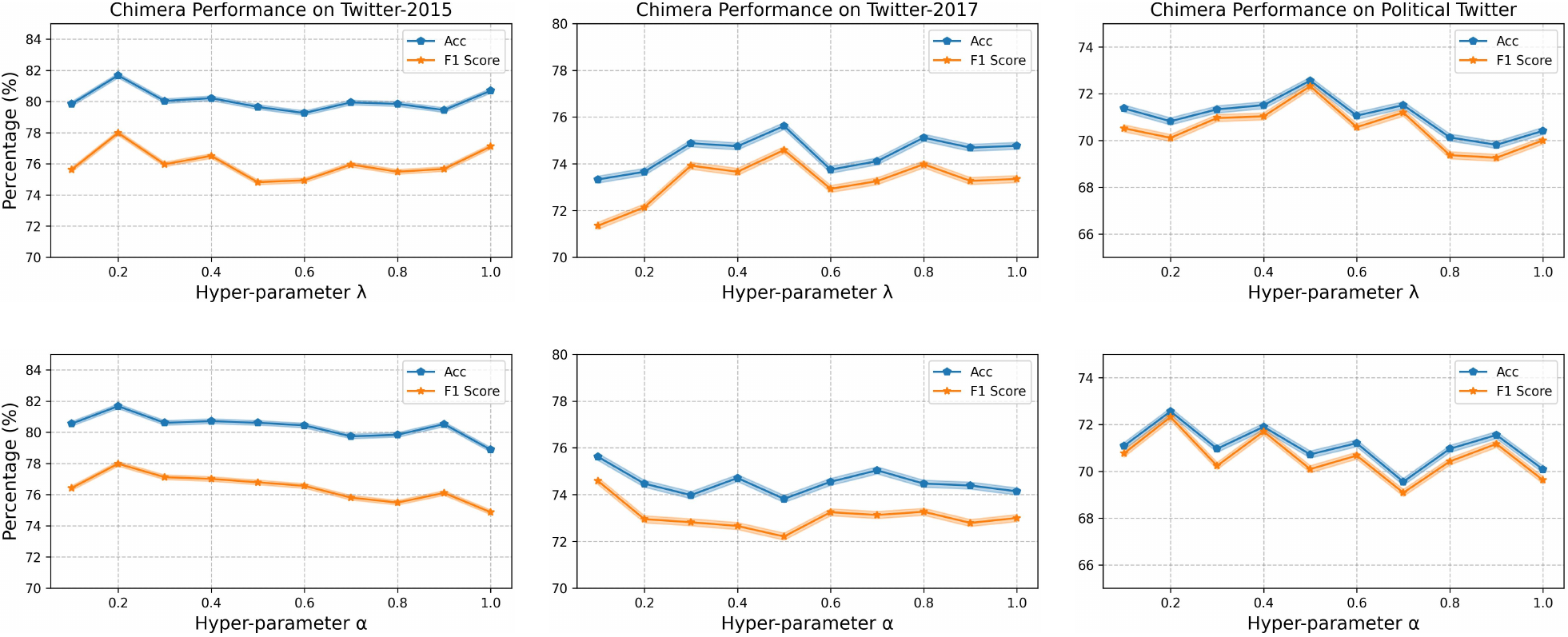}}
	\caption{Results ($\%$) on hyper-parameter of $\alpha$ and $\lambda$.}
	\label{fig:hyper}
\end{figure*}

\subsection{Ablation Study}
To systematically investigate the influence of the linguistic-aware semantic alignment module, including semantic and impression rationale reasoning as well as object-level fine-grained alignment, on sentiment prediction, we conducted a series of ablation studies and the results are shown in Table~\ref{exp:ablation}.

As presented in Table~\ref{exp:ablation}, the exclusion of semantic rationale ("w/o SRG") results in a noticeable performance decline across all three datasets. This effect is particularly pronounced on the Twitter-2017 and Political Twitter datasets, where nearly all evaluation metrics, including accuracy and F1 score, exhibit a reduction of approximately 2$\%$. Similarly, the absence of impression rationale reasoning ("w/o IRG") results in performance fluctuations on the Twitter-2015 and Political Twitter datasets. However, the most noticeable effect is observed on the Twitter-2017 dataset, where the model's performance exhibits a significant degradation, particularly in the sentiment classification task, with nearly a 4$\%$ drop in both accuracy and F1 score. 
\REO{The results ("w/o IRG \& AC") reveal consistent performance degradation in both Accuracy and F1-score across all three datasets. 
Particularly noteworthy is the model's inferior performance on Twitter-2017 and Political Twitter datasets compared to the baseline(w/o IRG). However, an unexpected performance improvement emerges in Twitter-2015, surpassing even the configuration retaining aesthetic captions as input.
This phenomenon may be attributed to dataset-specific characteristics in sample distribution. As detailed in Table 3, Twitter-2015 exhibits a significantly higher proportion of neutral-class samples compared to Twitter-2017 and Political Twitter.}
When the Chimera model is deprived of its reasoning abilities for both semantic and impression rationales (“w/o SRG $\&$ IRG”), its performance on sentiment classification declines to the lowest levels across all datasets. 
Specifically, a consistent reduction of approximately 4-5$\%$ is observed in nearly all metrics, underscoring the essential role of rationale-based reasoning in enhancing the effectiveness and accuracy of sentiment analysis tasks.
These results show that the influence of rationale reasoning differs across datasets. For Twitter-2017, with its balanced sentiment distribution (see Table~\ref{tab:twitter}), impression rationale has a greater impact on sentiment analysis. In contrast, both semantic and impression rationales contribute to the other two datasets, but neither is dominant.\\

The LSA branch plays a pivotal role in the Chimera model by bridging the semantic gap between textual and visual modalities, ensuring effective alignment of information across visual and textual data. Its removal (w/o LSA) consistently leads to a significant decline in performance across all datasets, as evident in the ablation study. For instance, on Twitter-2015, the accuracy drops from 81.61$\%$ to 80.54$\%$, and the F1 score decreases from 77.98$\%$ to 77.03$\%$. Similarly, for Twitter-2017, accuracy, and F1 score dropped to 73.72$\%$ and 70.96$\%$, respectively. By aligning linguistic and visual features, the branch allows the model to effectively interpret semantic overlaps and contrasts, enabling more accurate sentiment predictions. 

Object-level descriptions (e.g., facial expressions and object-level aesthetic captions) enrich the input sequence by providing object-level detailed visual context. The ablation study reveals that removing OD (w/o OD) causes noticeable performance drops. On Twitter-2015, accuracy drops by 1.65 percentage points, and the F1 score decreases by 1.90 percentage points. Similarly, on Twitter-2017, accuracy is reduced by 2.56 percentage points, while the F1 score drops by 3.74 percentage points. Without the OD, the model loses access to these fine-grained visual features, leading to diminished interpretability and accuracy, particularly in datasets where visual information plays a crucial role in determining sentiment. Additionally, the aesthetic caption is excluded from the input sequence to assess its impact on performance (w/o Aes-cap). As demonstrated in Table~\ref{exp:ablation}, the absence of aesthetic features results in a noteworthy decline in performance across all datasets, particularly in the impression rationale generation (IRG) task. This leads to Chimera exhibiting the poorest sentiment classification performance for IRG on the Twitter-2017 and Political Twitter datasets, which underscore the importance of aesthetic captions in guiding the model to generate coherent and emotionally nuanced impressions.

\subsection{Hyper-parameter Analysis}
We conduct a hyperparameter analysis to explore the impact of $\alpha$ and $\lambda$ on the Chimera model's performance across the Twitter-2015, Twitter-2017, and Political Twitter datasets. Hyperparameter $\alpha$ regulates the balance between sentiment classification (SC) and rationale generation components (semantic and impression rationales, SRG, and IRG), while $\lambda$ controls the weight of patch-token alignment within the overall loss function. As shown in Figure~\ref{fig:hyper}, for all datasets, a lower $\alpha$, which assigns greater weight to rationale generation, generally improves model performance, with values around 0.1 to 0.2 achieving the highest accuracy and F1 scores. This emphasizes the significance of integrating semantic and impression rationales in MASC. As $\alpha$ increases, favoring SC loss, performance plateaus or declines, particularly for the Political Twitter dataset, indicating that reduced emphasis on rationale generation diminishes the model's ability to capture fine-grained sentiment context effectively. Moreover, the results indicate that increasing $\lambda$ initially enhances model performance, with diminishing returns beyond a certain threshold. For the Twitter-2015 and Political Twitter datasets, moderate $\lambda$ values [0.2, 0.5] achieve optimal accuracy and F1 scores, while higher values ($\lambda > 0.6$) lead to performance stabilization or slight decline. This observation indicates that balanced alignment between visual and textual features enhances the model's interpretability and accuracy and excessively high $\lambda$ values may negatively impact performance, likely due to overemphasis on alignment at the expense of core sentiment classification. For Twitter-2017, a similar trend is observed, although performance variations are less pronounced.

\begin{table}[]
\caption{The evaluation resutls ($\%$) of rationale quality. The best-performing results highlighted in \textbf{bold}.}
  \resizebox{\columnwidth}{!}{
\begin{tabular}{@{}lcccccc@{}}
\toprule
\multirow{2}{*}{\textbf{Rationale Source}} & \multicolumn{2}{c}{\textbf{Twitter-2015}} & \multicolumn{2}{c}{\textbf{Twitter-2017}} & \multicolumn{2}{c}{\textbf{Political}} \\ \cmidrule(l){2-7} 
                    & \textbf{Acc}    & \textbf{F1}    & \textbf{Acc}    & \textbf{F1}    & \textbf{Acc}      & \textbf{F1}      \\ \midrule
\textbf{}               & \multicolumn{6}{c}{\textbf{Semantic Rationale}}                                          \\ \cmidrule(l){2-7} 
Ground-Truth              & \textbf{99.04}   & \textbf{99.04}   & \textbf{98.54}   & \textbf{98.54}   & \textbf{97.64}     & \textbf{97.64}    \\
Chimera                & 80.91        & 80.83       & 75.04        & 74.93       & 70.20         & 70.14         \\ \midrule
\textbf{}               & \multicolumn{6}{c}{\textbf{Impression Rationale}}                                          \\ \cmidrule(l){2-7} 
Ground-Truth              & \textbf{69.91}   & \textbf{69.90}   & \textbf{72.77}   & \textbf{72.71}   & \textbf{76.8}     & \textbf{76.87}    \\
Chimera                & 63.45        & 63.65       & 61.67        & 59.38       & 60.54         & 60.12         \\ \bottomrule
\end{tabular}
}
\label{tab:rationale_qua}
\end{table}

\begin{figure}[t!]
	\centerline{\includegraphics[width=1\columnwidth]{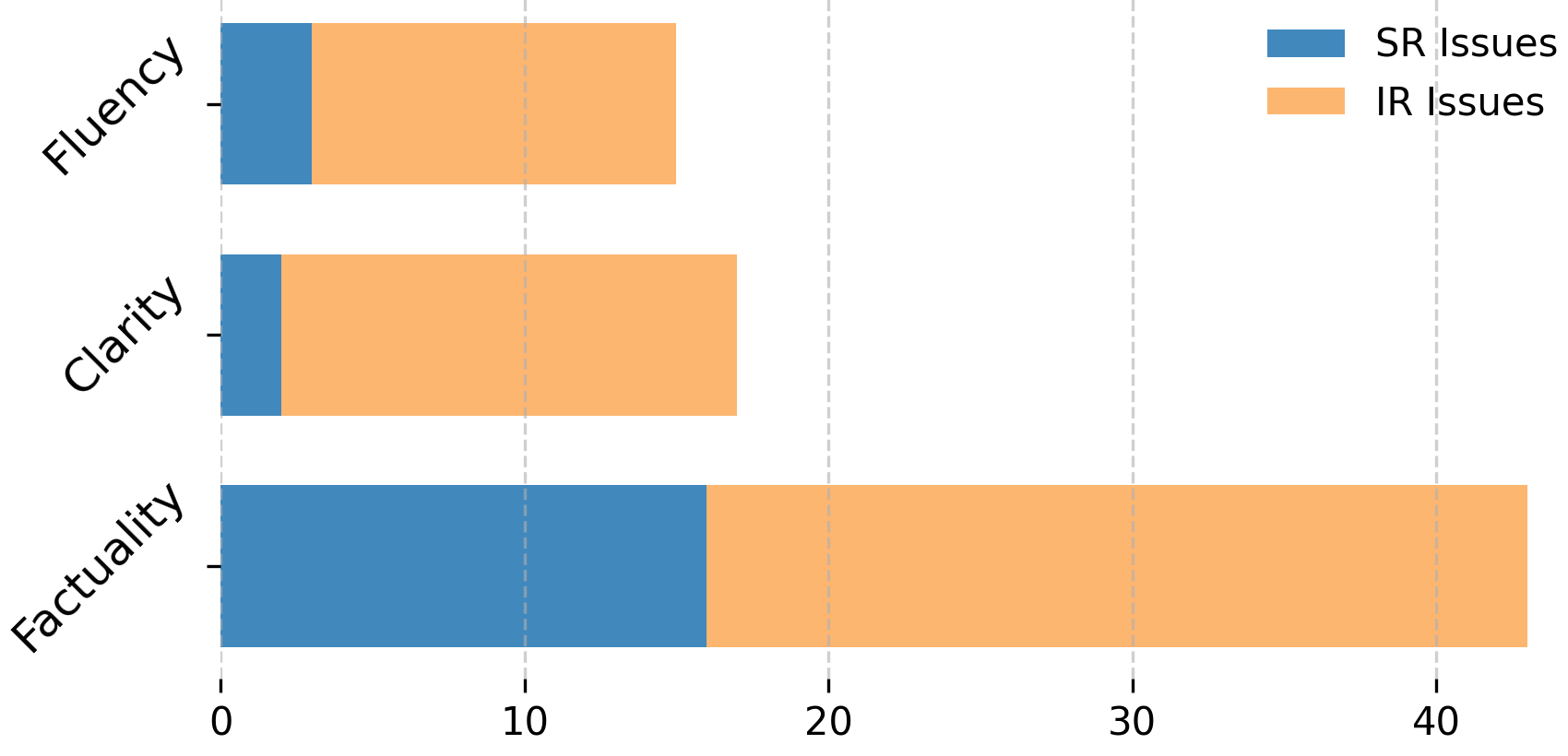}}
	\caption{Human evaluation of factuality, clarity and fluency for SR and IR.}
	\label{fig:manual_eval}
\end{figure}

\begin{figure*}[t!]
	\centerline{\includegraphics[width=1\textwidth]{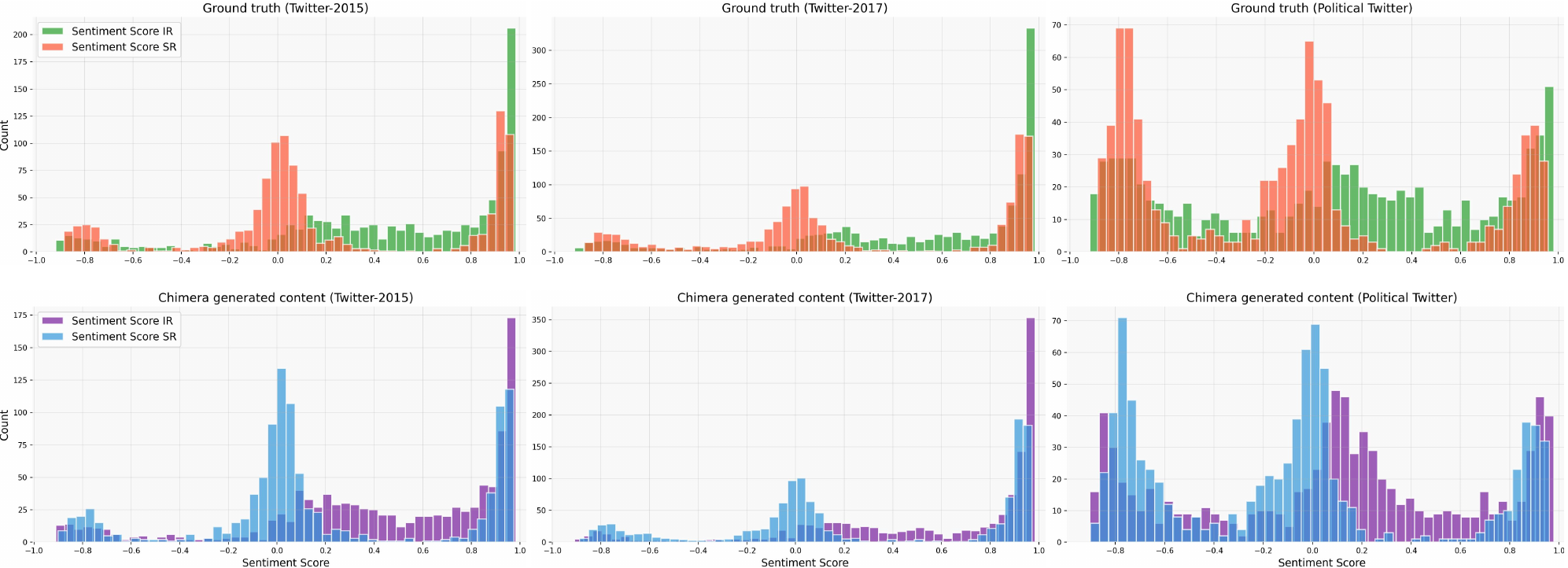}}
	\caption{Assessment of sentiment intensity for SR and IR in both ground truth data and Chimera-generated content.}
	\label{fig:quanti}
\end{figure*}
\section{In-depth Analysis}
\label{in:analysis}
\subsection{Quality Analysis of Rationale}
\REO{Table~\ref{tab:rationale_qua} provides an evaluation of the sentiment rationale quality for both the ground-truth and Chimera-generated content, aiming to analyze their impact on sentiment analysis. A pre-trained sentiment classification model~\cite{camacho2022tweetnlp} is employed to assess the intuitive sentiment quality of these rationales across three test datasets by inputting the rationales into the model and analyzing the sentiment predictions.
For both SR and IR, the results in the Ground-Truth row represent the upper performance bound. It is evident that the ground truth performance for SR significantly exceeds that of IR, indicating that semantic rationales are more critical for this task than impression rationales. }
\REO{We hypothesize that two factors contribute to this discrepancy. Firstly, as illustrated in Table~\ref{tab:twitter}, semantic rationales are shorter in length and straightforward, facilitating easy comprehension, while the emotions elicited by images are inherently more abstract and multifaceted. Secondly, the IR's reliance on visual cues contrasts sharply with the Twitter dataset's text-centric sentiment distribution. }
\REO{
Prior research has shown that a considerable majority of targets (around 58\%) are absent from images~\cite{yu2022targeted}, and most targets (93\% in Twitter-2015) exhibit emotional coherence with their textual counterparts~\cite{ye2023rethinkingtmsc}. This misalignment underscores the dataset's limitations in evaluating IRs and necessitates a nuanced understanding of the interplay between visual and textual sentiment representations.
}

\REO{
A total of 180 samples were randomly selected for human evaluation, with 100 samples drawn from the training set, 40 from the testing set, and 40 from the validation set of both the Twitter-2015 and Twitter-2017 datasets. Four native English speakers with Master's degrees in the arts were recruited to assess the quality of the rationale data based on three criteria: (1) factuality, evaluating whether the rationale is grounded in accurate and verifiable information; (2) clarity, assessing the logical structure and comprehensibility of the rationale; and (3) fluency, measuring the grammatical accuracy and smoothness of the language used. The Fleiss' Kappa ($\kappa$) values for the initial evaluation across the four raters were as follows: factuality $\kappa = 0.922$, clarity $\kappa = 0.945$, and fluency $\kappa = 0.960$. In cases of disagreement, the evaluators engaged in discussions to reach a consensus.
}
\REO{
Figure~\ref{fig:manual_eval} presents the results of the human evaluation. It can be observed that SR consistently exhibits higher quality across all metrics, which verifies that the employed LLM is capable of generating appropriate rationale data for specific tasks when provided with concrete ground-truth labels. In comparison to SR, IR demands a more in-depth understanding of visual content and is inherently more subjective. Consequently, IR is more prone to issues of factuality and clarity, as interpreting the abstract aesthetic and emotional elements conveyed by an image often involves subjective reasoning, which may lead to misalignment with objective ground truths or human expectations.
}

\subsection{Quantitative Analysis of Rationale}

We conduct a quantitative analysis on the test sets of ground truth and Chimera-generated content to examine the impact of varying levels of sentiment intensity in cognitive rationales on the accuracy of sentiment prediction, including their potential to amplify or diminish predictive performance. 
As illustrated in Figure~\ref{fig:quanti}, the sentiment intensity distributions of Twitter-2015 and Twitter-2017 reveal distinct patterns. Specifically, the sentiment intensity of IR demonstrates a noticeable bias toward positive values, whereas the sentiment intensity of SR aligns more closely with the sentiment polarity label distribution presented in Table~\ref{tab:twitter}. 

This observation suggests that IR demonstrates a bias toward positive samples, increasing the model's confidence in predicting positive instances. While this bias may be beneficial for datasets with a higher proportion of positive samples (e.g., Twitter-2017), it could lead to additional bias in datasets with a limited representation of positive samples. This finding is further corroborated by the ablation study results, which reveal that the performance of the Chimera model without IR is worse on Twitter-2017 compared to its performance on Twitter-2015. Another notable observation is that, for the ground truth of the Political Twitter dataset, the sentiment intensity distribution of IR is relatively uniform across all ranges. In contrast, the Chimera-generated content for IR exhibits a more distinguishable sentiment intensity distribution compared to the ground truth, which further validates the quality of SR, the effectiveness of the proposed Chimera training paradigm, and the robustness of Chimera's performance.

\begin{figure}[t!]
	\centerline{\includegraphics[width=1\columnwidth]{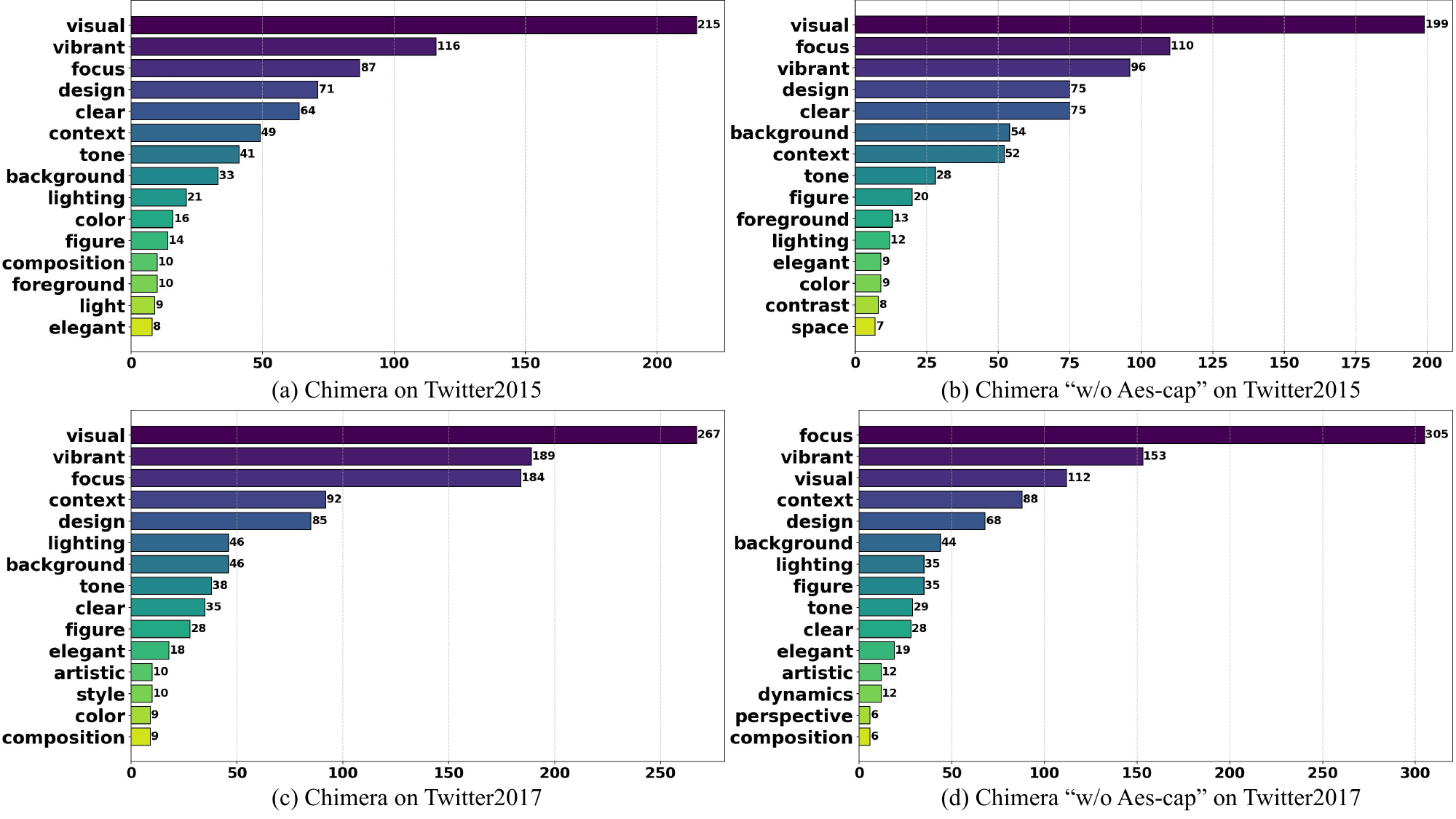}}
	\caption{Visualization of the top 15 most frequent aesthetic-related words in generated IR.}
	\label{fig:aes_count}
\end{figure}

\begin{figure*}[t!]
	\centerline{\includegraphics[width=0.95\textwidth]{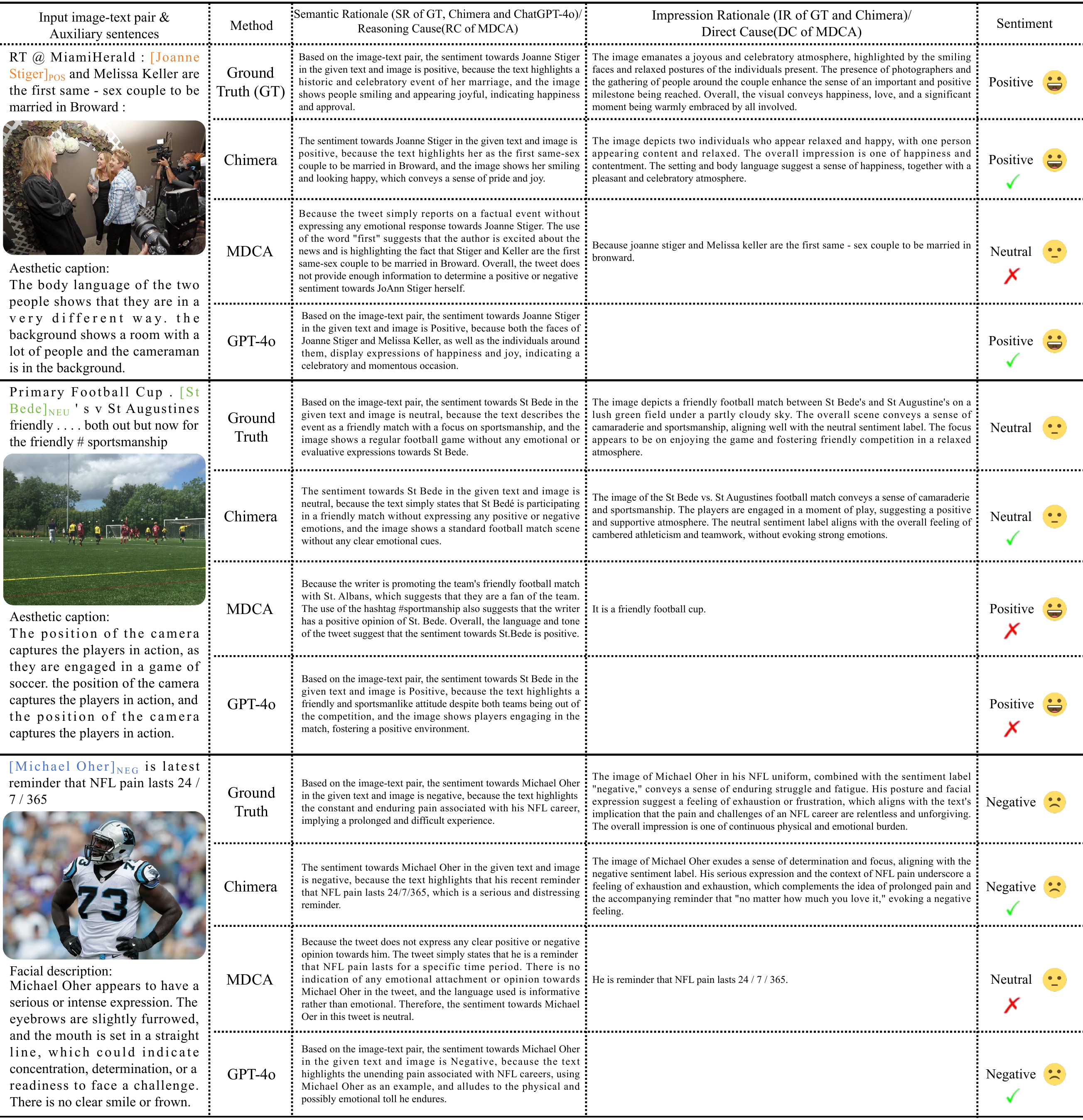}}
	\caption{Three examples showcasing the predictions generated by Chimera, MDCA, and GPT-4o are presented for analysis. During the evaluation process, GPT-4o exclusively produces the semantic rationale (SR). The input image-text pair and auxiliary sentences are utilized solely by Chimera. For MDCA, the reasoning cause (RC), direct cause (DC), and sentiment prediction are derived through direct inference.}
	\label{fig:case}
\end{figure*}

\subsection{Impact of Aesthetic Attributes on Sentiment}

To investigate the impact of image aesthetic attributes on sentiment analysis, we visualize the frequency of aesthetic-related words within the impression rationales generated by our proposed Chimera model and its variant ``Chimera w/o Aes-cap'' on the Twitter-2015 and Twitter-2017 test sets. Specifically, we visualize the top 15 most frequent aesthetic-related words within the generated IR, based on the aesthetic attributes defined by Milena~\etal~\cite{ivanova2020aesthetics}. As shown in Figure~\ref{fig:aes_count}, the frequency analysis of aesthetic-related words for Chimera on Twitter-2015 and Twitter-2017 reveals that ``visual,'' ``vibrant,'' ``focus,'' and ``design'' prominently appear across both datasets. These terms, associated with visual clarity, expressive quality, image composition, and cohesiveness, align with the model's improved accuracy and F1 scores. However, excluding the aesthetic caption from the input results in subtle shifts in the frequency distribution of these aesthetic-related terms. For Twitter-2015, the overall frequency distribution of aesthetic-related terms shows minimal change, with a slight increase in ``focus'' and a decrease in ``vibrant''. In contrast, for Twitter-2017, ``focus'' shows a significant increase, while ``visual'' and ``vibrant'' decrease notably. Combined with the ablation study results in Table~\ref{exp:ablation}, the absence of aesthetic captions in the input leads to the worst sentiment analysis performance across all datasets on IRG. This highlights the critical role of aesthetic captions in enhancing the model's understanding of image aesthetics, particularly in datasets like Twitter-2017 with balanced sentiment distributions. Specifically, attributes such as ``visual'' and ``vibrant'' positively contribute to sentiment analysis performance, whereas ``focus'' appears to significantly impair it. We speculate that since "focus" emphasizes specific image elements, potentially leads to an unbalanced interpretation of visual content. This localized emphasis can narrow the model's analytical scope, prioritizing details at the expense of broader context and compositional harmony. Consequently, the model may struggle to capture holistic aesthetic and emotional cues essential for accurate sentiment classification.

\begin{table}[!thb]
\caption{The experimental results ($\%$) of GPT-4o on the MASC task under a zero-shot setting are presented. The best-performing results highlighted in \textbf{bold}. The term “dis” refers to the percentage of samples where the sentiment polarity associated with a specific aspect cannot be discerned.}
\resizebox{\columnwidth}{!}{
\begin{tabular}{@{}lcccccc@{}}
\toprule
\multirow{2}{*}{\textbf{Method}} & \multicolumn{3}{c}{\textbf{Twitter-2015}}              & \multicolumn{3}{c}{\textbf{Twitter-2017}}              \\ \cmidrule(l){2-7} 
                 & \textbf{Acc}     & \textbf{F1}     & \textbf{Dis}     & \textbf{Acc}     & \textbf{F1}     & \textbf{Dis}     \\ \midrule
Chimera             & \textbf{81.61}    & \textbf{77.98}    & -          & \textbf{75.62}    & \textbf{74.59}    & -          \\ \midrule
GPT-4o            & 46.87        & 47.47        & 0.2         & 56.08        & 53.28        & 0.5         \\
GPT-4o w/o image       & 67.02        & 62.38        & -          & 59.64        & 60.35        & -          \\ \midrule
                 & \multicolumn{1}{l}{} & \multicolumn{1}{l}{} & \multicolumn{1}{l}{} & \multicolumn{1}{l}{} & \multicolumn{1}{l}{} & \multicolumn{1}{l}{}
\end{tabular}
}\label{exp:LLMs}
\end{table}
\subsection{Comparison with Large Language Models}

We evaluate the performance of GPT-4o on the MASC task under a zero-shot setting. As shown in Table~\ref{exp:LLMs}, GPT-4o achieves an accuracy of 46.87$\%$ and an F1 score of 47.47$\%$, which is substantially lower than Chimera, which reports 81.61$\%$ accuracy and 77.98$\%$ F1 score. On the Twitter-2017 dataset, GPT-4o shows an improvement with an accuracy of 56.08$\%$ and an F1 score of 53.28$\%$. However, this performance still trails behind Chimera, which reports 75.62$\%$ accuracy and 74.59$\%$ F1 score. Surprisingly, removing the image input results in an improvement in the model's accuracy and F1 score, reaching 67.02$\%$ and 62.38$\%$ on the Twitter-2015 dataset, respectively. This observation contrasts sharply with the phenomenon observed in the baseline model. Similarly, in the Twitter-2017 dataset, the performance of GPT-4o without image input is slightly better than with the image input. We speculate that in task-specific models, incorporating image data typically improves sentiment classification performance, as these models are fine-tuned to leverage multi-modal inputs effectively. However, in a zero-shot setting, GPT-4o operates based on its general pre-trained knowledge, which may not be fully optimized for combining textual and visual inputs for sentiment analysis. In this setting, adding image input may introduce noise rather than meaningful information. Moreover, GPT-4o has a low Dis value on both datasets, which slightly decreases to 0 when the image input is removed. This further suggests that the model's ability to distinguish sentiment polarity is, to a certain extent, influenced by the inclusion of the visual modality.

\subsection{Case Study}
An additional case study is performed to provide a more comprehensive evaluation of the effectiveness of the proposed Chimera model. 
Figure~\ref{fig:case} illustrates three representative examples, each corresponding to positive, neutral, and negative samples, respectively. As illustrated in the first example, MDCA is the sole model to predict ``Neutral'' for the target ``Joanne Stiger,'' whereas the other three models accurately predict ``Positive''. This result is primarily due to the RC and DC generated by MDCA, which lack the expression of positive or negative sentiment. Notably, the RC predominantly emphasizes the textual content, overlooking the joyful atmosphere conveyed through the image. In the second example, an intriguing observation is that the situation is the exact opposite of the previous case. \REO{Here, only Chimera correctly predicts the sentiment polarity of the specific target, ``St. Bede'' as ``Neutral'' whereas both GPT-4o and MDCA incorrectly classify it as ``Positive''. It is observed that the SR of GPT-4o and the RC of MDCA both convey a positive sentiment, largely due to an overinterpretation and extrapolation of the textual content. In contrast, Chimera demonstrates accurate prediction by appropriately integrating a balanced understanding of the image content and its aesthetic attributes. In the final example, both Chimera and GPT-4o accurately identify the sentiment polarity of ``Michael Oher'' as ``Negative''.} MDCA's incorrect prediction of "Neutral" may be attributed to its generated RC and DC failing to account for the individual's expression, thereby overlooking critical semantic cues present in the visual content. With the aid of facial descriptions, Chimera effectively captures and aligns fine-grained emotional cues from visual content, enabling it to generate coherent SR and IR and achieve accurate predictions. The above representative instances further verify that incorporating cognitive and aesthetic sentiment causality enhances sentiment classification accuracy in MABSA.

\section{Conclusion}\label{sec:conclusion}

In this paper, we propose a cognitive sentiment causality understanding framework tailored for multimodal aspect-based sentiment classification. The framework, which is novel in its approach, consists of four primary components: linguistic-aware semantic alignment, a translation module, rationale dataset construction, and rationale-aware learning. The linguistic-aware semantic alignment component facilitates visual patch-token level alignment through dynamic patch selection and semantic patch calibration. The translation module transforms holistic image and object-level visual information into corresponding emotion-laden textual representations. The rationale dataset construction involves designing refined prompts and leveraging LLMs to generate semantic and impression rationale. Finally, rationale-aware learning incorporates semantic explanations and affective-cognitive resonance to enhance the model's capacity to understand cognitive sentiment causality. Experimental results on three Twitter datasets demonstrate that the proposed Chimera achieves performance gains over SOTA baselines.

\section*{Acknowledgments}
This research is supported by the Shanghai Science and Technology Innovation Action Plan (No. 24YF2710100), the Shanghai Special Project to Promote High-quality Industrial Development (No. RZ-CYAI-01-24-0288), the National Nature Science Foundation of China (No. 62477010), the Science and Technology Commission of Shanghai Municipality Grant (No. 22511105901, No. 21511100402), the Ministry of Education, Singapore under its MOE Academic Research Fund Tier 2 (STEM RIE2025 Award MOE-T2EP20123-0005) and by the RIE2025 Industry Alignment Fund – Industry Collaboration Projects (IAF-ICP) (Award I2301E0026), administered by A*STAR, as well as supported by Alibaba Group and NTU Singapore. 

\bibliographystyle{IEEEtran}
\bibliography{reference}
\vspace{-10 mm}
\begin{IEEEbiography}[{\includegraphics[width=1in,height=1.25in,clip,keepaspectratio]{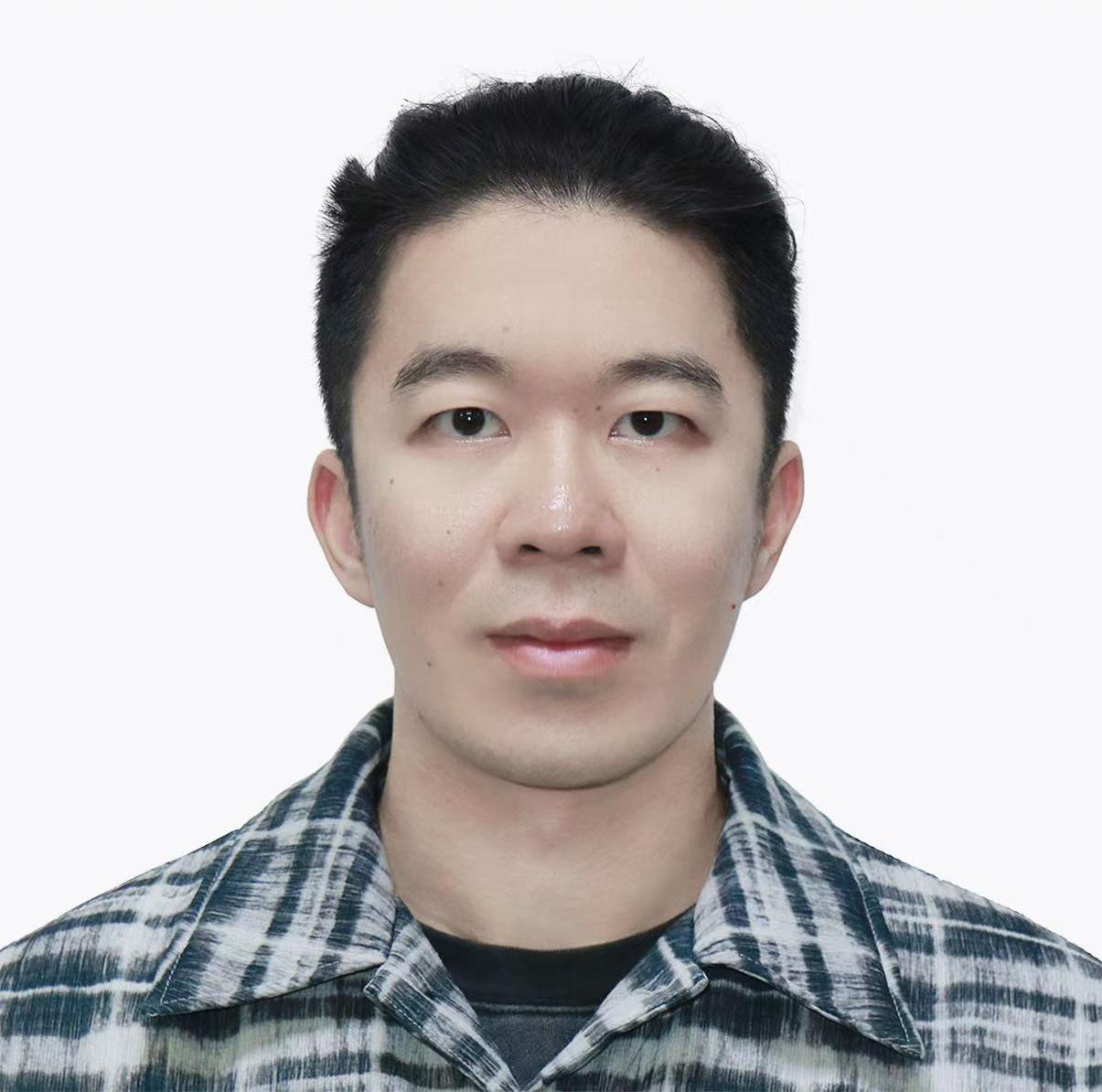}}]{Luwei Xiao}{\,} is currently pursuing his Ph.D. degree in the School of Computer Science and Technology at East China Normal University, Shanghai, China, under the supervision of Prof.~Liang He. He is presently conducting an academic visit to the College of Computing and Data Science at Nanyang Technological University, Singapore, under the supervision of Prof.~Erik Cambria, with funding support from the China Scholarship Council (CSC). His research interests encompass multimodal learning, sentiment analysis, and image aesthetic assessment.
\end{IEEEbiography}

\begin{IEEEbiography}[{\includegraphics[width=1in,height=1.25in,clip,keepaspectratio]{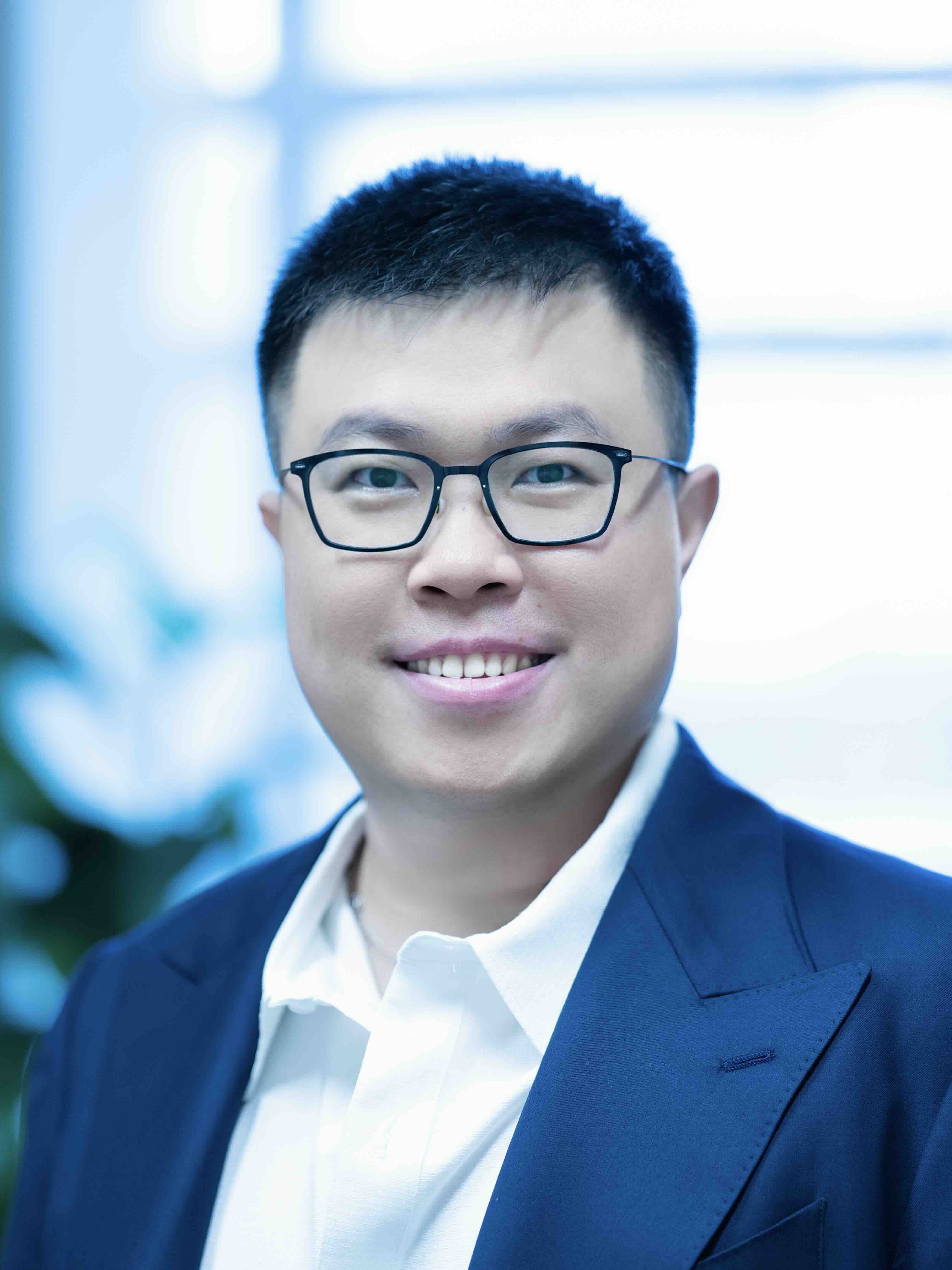}}]{Rui Mao}{\,}is a Research Scientist and Lead Investigator at Nanyang Technological University. He obtained his Ph.D. degree in Computing Science from the University of Aberdeen. His research interest lies in NLP, cognitive computing, and their applications in finance and cognitive science. He and his funded company (Ruimao Tech) have developed an end-to-end system (MetaPro) for computational metaphor processing and a neural search engine (wensousou.com) for searching Chinese ancient poems with modern language. He served as Area Chair in COLING and EMNLP and Associate Editor in IEEE Transactions on Affective Computing, Expert Systems, Information Fusion and Neurocomputing. Contact him at \href{mailto:rui.mao@ntu.edu.sg}{rui.mao@ntu.edu.sg}.
\end{IEEEbiography}

\begin{IEEEbiography}[{\includegraphics[width=1in,height=1.25in,clip,keepaspectratio]{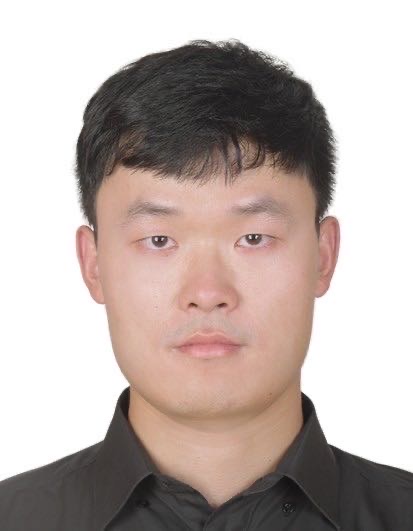}}]{Shuai Zhao} obtained his Ph.D. degree from Jinan University in 2024. He spent one year as a visiting student and six months as a research assistant at the School of Computer Science and Engineering, Nanyang Technological University. He is now a Postdoctoral Researcher at the College of Computing and Data Science, Nanyang Technological University. His current research interests include deep learning and natural language processing for code generation, summary generation, text classification and backdoor attacks.
\end{IEEEbiography}

\begin{IEEEbiography}[{\includegraphics[width=1in,height=1.25in,clip,keepaspectratio]{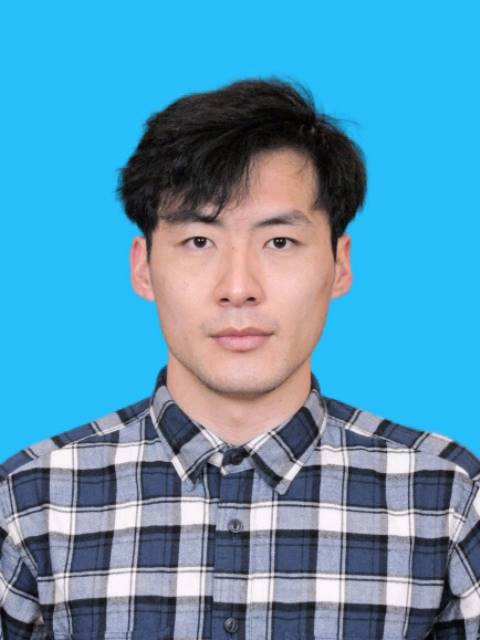}}]{Qika Lin} received his Ph.D. degree at Xi'an Jiaotong University. Currently, he is a Research Fellow at the National University of Singapore. His research interests include natural language processing, knowledge reasoning, and multimodal learning.
He has published papers in top-tier journals/conferences, including TKDE, ACL, SIGIR, KDD, ICDE, and IJCAI.
He has actively contributed to several journals/conferences as a reviewer or PC member, including TPAMI, IJCV, TKDE, TMC, TNNLS, NeurIPS, ICLR, SIGIR, ACL, and EMNLP.
He also served as a Guest Editor of IEEE TCSS and Information Fusion.
\end{IEEEbiography}

\begin{IEEEbiography}[{\includegraphics[width=1in,height=1.25in,clip,keepaspectratio]{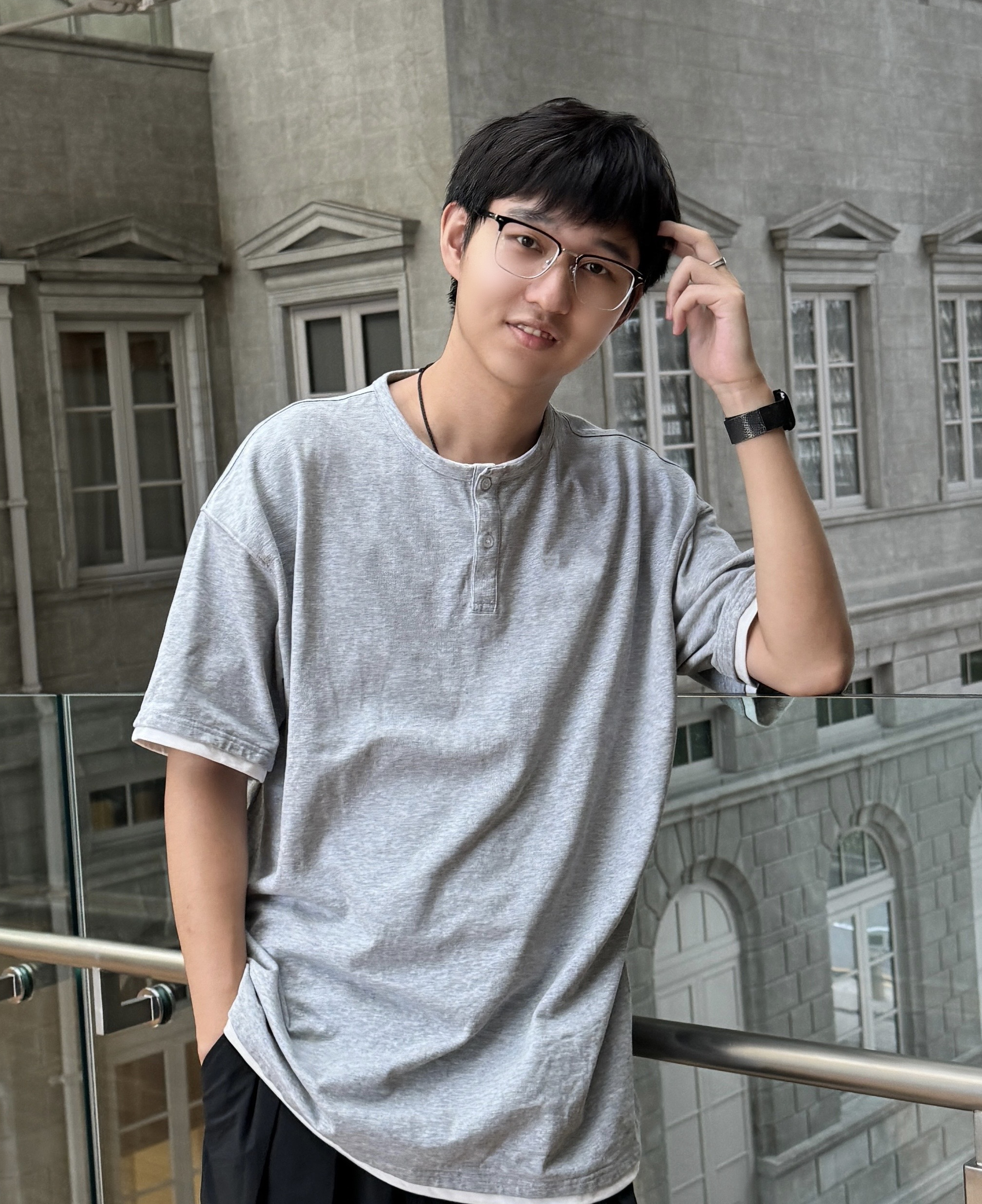}}]{Yanhao Jia}{\,} is a phd student at Nanyang Technological University. He obtained his bechealor degree in Computing Science from Shandong University. He has published over seven conference/journal papers on ECCV/NeurIPS/IEEE Trans on nuclear science and been the reviewer for ACM MM and ECCV.
\end{IEEEbiography}

\begin{IEEEbiography}[{\includegraphics[width=1in,height=1.25in,clip,keepaspectratio]{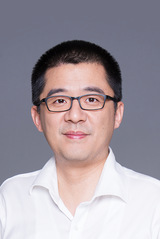}}]{Liang He}{\,} received his PhD degree from the Department of Computer Science and Technology, East China Normal University, China. He is now a professor and the Vice Dean of the School of Computer Science and Technology, East China Normal University. His current research interest includes Natural Language Processing, Knowledge Processing, and Human in the Loop for Decision-making.
\end{IEEEbiography}

\begin{IEEEbiography}[{\includegraphics[width=1in,height=1.25in,clip,keepaspectratio]{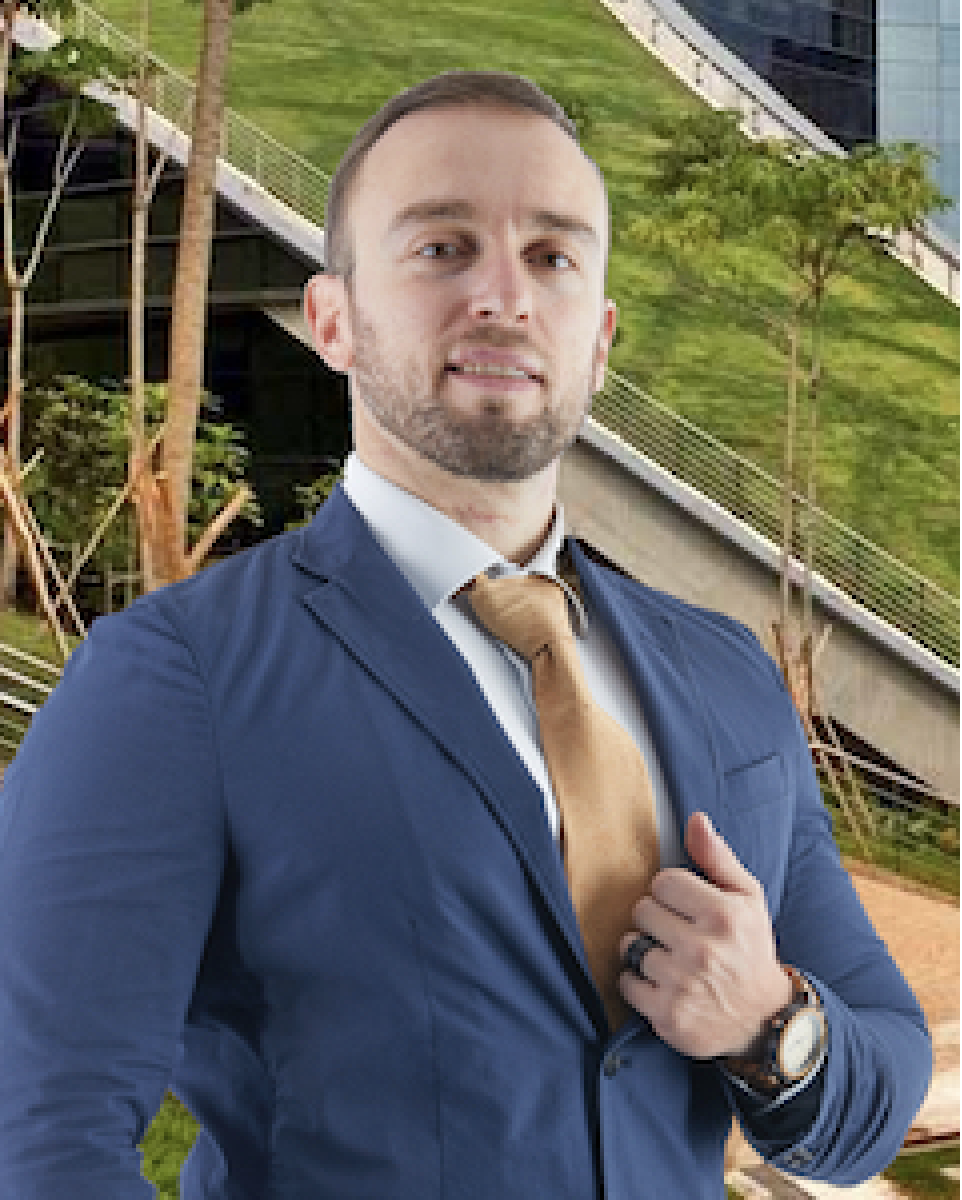}}]{Erik Cambria}{\,}is a Professor at Nanyang Technological University, where he also holds the appointment of Provost Chair in Computer Science and Engineering, and Founder of several AI companies, such as SenticNet, offering B2B sentiment analysis services, and finaXai, providing fully explainable financial insights. His research focuses on neurosymbolic AI for interpretable, trustworthy, and explainable affective computing in domains like social media monitoring, financial forecasting, and AI for social good. He is an IEEE Fellow, Associate Editor of various top-tier AI journals, e.g., Information Fusion and IEEE Transactions on Affective Computing, and is involved in several international conferences as keynote speaker, program chair and committee member. Contact him at \href{mailto:cambria@ntu.edu.sg}{cambria@ntu.edu.sg}.
\end{IEEEbiography}

\end{document}